\documentclass[lettersize,journal]{IEEEtran}
\usepackage{amsmath,amsfonts}
\usepackage{algpseudocode}
\usepackage{array}
\usepackage[caption=false,font=normalsize,labelfont=sf,textfont=sf]{subfig}
\usepackage{textcomp}
\usepackage{stfloats}
\usepackage{url}
\usepackage{verbatim}
\usepackage{graphicx}
\usepackage{cite}
\usepackage[hidelinks]{hyperref}
\hypersetup{
    colorlinks=false,   
    linktoc=all,        
    linkcolor=black,     
}
\usepackage{booktabs}
\usepackage{wrapfig}
\usepackage[linesnumbered,ruled,lined,boxed]{algorithm2e}
\usepackage[numbers,sort&compress]{natbib}
\usepackage{url}

\begin{document}

\title{Hard Sample Mining Enabled Supervised Contrastive Feature Learning for Wind Turbine Pitch System Fault Diagnosis}


\author{Zixuan Wang, Bo Qin, Mengxuan Li, Chenlu Zhan, Mark D.\ Butala, Peng Peng,~\IEEEmembership{Member,~IEEE}, and Hongwei Wang,~\IEEEmembership{Member,~IEEE}
\thanks{Zixuan Wang and Bo Qin contribute equally to this work.}
\thanks{Zixuan Wang is with the College of Biomedical Engineering and Instrument Science in Zhejiang University, Hangzhou, 310013, China. (E-mail: \href{mailto:zixuanw.20@intl.zju.edu.cn}{zixuanw.20@intl.zju.edu.cn}.}
\thanks{Mengxuan Li and Chenlu Zhan are with the College of Computer Science and Technology in Zhejiang University, Hangzhou, 310013, China. (E-mail: \href{mailto:mengxuanli@intl.zju.edu.cn}{mengxuanli@intl.zju.edu.cn}).}
\thanks{Bo Qin, Mark D.\ Butala, Peng Peng and Hongwei Wang are with Zhejiang University and the University of Illinois Urbana–Champaign Institute, Haining, 314400, China. (E-mail:  \href{mailto:bo.20@intl.zju.edu.cn}{bo.20@intl.zju.edu.cn}, 
\href{mailto:markbutala@intl.zju.edu.cn}{markbutala@intl.zju.edu.cn}, \href{mailto:pengpeng@intl.zju.edu.cn}{pengpeng@intl.zju.edu.cn}, \href{mailto:hongweiwang@intl.zju.edu.cn}{hongweiwang@intl.zju.edu.cn}).}
\thanks{Corresponding authors: Peng Peng and Hongwei Wang}
        }



\maketitle

\begin{abstract}
The efficient utilization of wind power by wind turbines relies on the ability of their pitch systems to adjust blade pitch angles in response to varying wind speeds. However, the presence of multiple health conditions in the pitch system due to the long-term wear and tear poses challenges in accurately classifying them, thus increasing the maintenance cost of wind turbines or even damaging them. This paper proposes a novel method based on hard sample mining-enabled supervised contrastive learning (HSMSCL) to address this problem. The proposed method employs cosine similarity to identify hard samples and subsequently, leverages supervised contrastive learning to learn more discriminative representations by constructing hard sample pairs. 
Furthermore, the hard sample mining framework in the proposed method also constructs hard samples with learned representations to make the training process of the multilayer perceptron (MLP) more challenging and make it a more effective classifier.
The proposed approach progressively improves the fault diagnosis model by introducing hard samples in the SCL and MLP phases, thus enhancing its performance in complex multi-class fault diagnosis tasks.

To evaluate the effectiveness of the proposed method, two real datasets comprising wind turbine pitch system cog belt fracture data are utilized. The fault diagnosis performance of the proposed method is compared against existing methods, and the results demonstrate its superior performance. The proposed approach exhibits significant improvements in fault diagnosis performance, providing promising prospects for enhancing the reliability and efficiency of wind turbine pitch system fault diagnosis.

\end{abstract}

\begin{IEEEkeywords}
Wind turbine, Pitch system, Fault diagnosis, Hard sample mining, Class imbalance
\end{IEEEkeywords}

\section{Introduction}
\IEEEPARstart{W}{ind} energy has been widely used for decades. According to the Global Wind Energy Council (GWEC) statistic in 2022, nearly 94 GW of new wind turbine capacity was installed in 2021, bringing the global wind power capacity to 837 GW \cite{gwec2022global}. The maintenance cost is also increasing with the increasing number of wind turbines. Therefore, condition monitoring and fault diagnosis for wind turbines is crucial to control the operating costs of wind turbines. 

A wind turbine is a complex electromechanical system consisting of blades, rotors, pitch system, yaw system, gearbox, generator, and other components \cite{qiao2015survey}. According to a survey conducted by the China Wind Energy Association (CWEA) between 2010 and 2012 among 47 wind turbine manufacturers, component suppliers and developers, damage to these components is the main cause of wind turbine failures. Among them, the frequency of failure of the pitch system ranked high \cite{lin2016fault}. 

\begin{figure}[!t]
\centering
\includegraphics[width=0.95\linewidth]{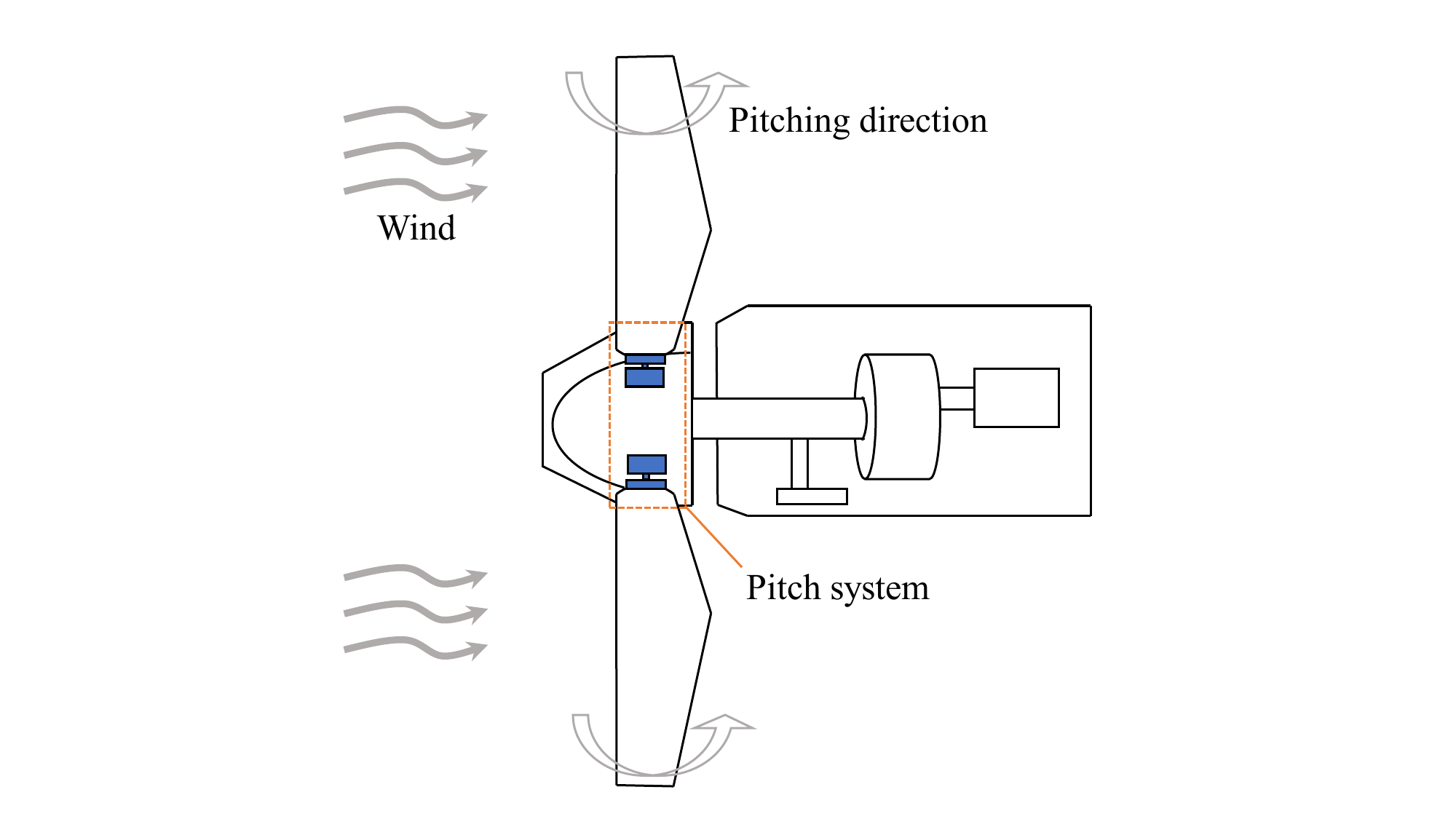}
\caption{The general structure of the wind turbine pitch system}
\label{pitchstructure}
\end{figure}

The general structure of the wind turbine pitch system is shown in Fig.\ref{pitchstructure}. The Pitch system can adjust the blade pitch angle in real-time according to the change of wind speed, so as to make full use of wind energy for power generation and reduce the impact of wind on the blade \cite{perez2013wind}. In addition, the pitch system can feather the blade to achieve a safe shutdown when the wind speed is too high or during an emergency shutdown \cite{hansen2004review}. There are two types of pitch systems for wind turbines: electric motor drive and hydraulic drive. Due to the oil leakage and maintenance problems of hydraulic drive, the pitch systems of wind turbines in China are mainly driven by electric motors \cite {lin2016fault}. The main failures of electric motor-driven pitch systems include electric motor failure, transmission component failure, and control system failure. Therefore, the timely diagnosis of wind turbine pitch system faults can effectively reduce downtime and losses. 

The methods of wind turbine fault diagnosis are mainly divided into signal analysis-based methods, model-based methods, and data-driven methods \cite{helbing2018deep,li2021wind,habibi2019reliability}.
Signal analysis-based methods acquire data through sensors installed in various components of the wind turbine and extract fault-related features for fault diagnosis \cite{wang2019vibration,liu2020review,salameh2018gearbox}.  For example, \citet{an2012application} put forward a fault diagnosis method of wind turbine bearing based on intrinsic time-scale decomposition. \citet{zhang2018wavelet} proposed a wavelet energy transmissibility function that is robust to noise and sensitive to system property changes to diagnose the fault in wind turbine bearings. \citet{tang2010wind} proposed an effective wind turbine fault diagnosis method with good denoising performance based on the Morlet wavelet transformation and Wigner-Ville distribution. Although signal analysis-based methods possess high fault detection accuracy, they are extremely dependent on additional sensors, which themselves will inevitably fail, thus affecting the accurate diagnosis of faults. In addition, the arrangement of sensors also increases the operational cost and wiring complexity of the wind turbine \cite{qiao2015survey2}. 

Model-based approaches construct a quantitative or qualitative model for fault diagnosis through prior knowledge of the operating modes of the wind turbine system \cite{abichou2014fault}. For example, \citet{Chaaban2014} modeled a utility-scale turbine using an aero-hydro-servo-elastic simulation tool to simulate the faults inside the pitch system and to study their effects as a function of the fault magnitude and wind speed. \citet{odgaard2012model} proposed a fault-tolerant observer scheme to provide estimates of various components for a wind turbine. \citet{lan2018fault} proposed an adaptive sliding observer to estimate wind turbine parametric pitch actuator faults. Model-based wind turbine fault diagnosis techniques do not rely on additional hardware components, but their diagnostic performance depends entirely on the mathematical model. The complex structure of a wind turbine and its dynamic operating states make it particularly challenging to construct a suitable mathematical model \cite{9774410}. 

Data-driven methods use a large amount of historical data which can be obtained through the supervisory control and data acquisition (SCADA) system installed in the wind turbine to train the fault diagnosis model. For example, \citet{gao2018novel} proposed a wind turbine fault diagnosis method based on integral extension load mean decomposition multiscale entropy and least squares support vector machine.  \citet{lei2019fault} proposed an end-to-end Long Short-term Memory (LSTM) model for fault diagnosis of wind turbines. \citet{pashazadeh2018data} proposed a data-driven approach based on the fusion of several classifiers for fault detection and isolation in wind turbines. However, data-driven approaches still face some challenges. (1). The complex and variable operating states of wind turbines lead to a huge amount of data collected by SCADA systems with the characteristics of class imbalance, nonlinearity, and nonstationarity, which makes it difficult to effectively extract features from the data; (2). Existing studies rarely focus on the problem of hard samples in multiple health states which is difficult to distinguish in wind turbine fault diagnosis tasks.



This paper proposes a novel fault diagnosis method for wind turbine pitch systems that combines supervised contrastive learning and hard sample mining. The purpose of the proposed method is to better distinguish the faults in the wind turbine pitch system whose severity varies with time. In order to achieve this purpose, we face two challenges. First, the amount of SCADA data on wind turbines is extremely large, so how to extract discriminative features from SCADA data is the problem we face. Second, there are many different health conditions in the wind turbine pitch system, which may be difficult to distinguish between them due to the high degree of similarity, so how to accurately distinguish between multiple health conditions is another challenge. These two challenges are the key issues in building a fault diagnosis model for wind turbine pitch systems based on SCADA data.

The main contributions of this paper can be summarized as follows.
\begin{itemize}
    \item A two-step approach consisting of supervised contrastive learning and a multi-layer perceptron is used to diagnose faults in wind turbine pitch systems. Supervised contrastive feature learning extracts discriminative features from large amounts of SCADA data. The extracted features are then utilized to classify the data and discriminate faults by the multi-layer perceptron.
    \item A hard sample mining framework based on cosine similarity is used to distinguish the hard samples in wind turbine SCADA data with multiple health conditions. To the best of our knowledge, few similar studies have been conducted in the field of wind turbine pitch system fault diagnosis, and the hard sample mining framework proposed in this paper fills the gap in this field.
    \item The proposed hard sample mining method is a unified framework acting on both supervised contrastive learning and multi-layer perceptron.
\end{itemize}

 The rest of this paper is organized as follows. First, the background theory is discussed in Section II. Then, the proposed HSMSCL is described in detail in Section III. The data used for the experiments and the analysis of the experimental results are presented in Section IV. Finally, the conclusions of this paper are given in Section V.

\section{Background Theory}

\subsection{Related works}
In practice, the wind turbine pitch system operates in a normal state most of the time, while the wear and tear of the pitch system is a gradual process, with multiple health states before complete damage. Therefore, the fault diagnosis problem of the pitch system is a typical imbalanced multi-classification problem. The common methods for solving this problem mainly include data-level methods, algorithm-level methods, and normal behavior model-based(NBM) methods \cite{tanha2020boosting,tautz2017using}. Data-level methods make the data balanced by resampling them, including the undersampling of the major class and the oversampling of the minor class \cite{yang2023fault}. For example, \citet{zhang2018imbalanced} proposed a novel weighted minority oversampling approach for imbalanced data problems in real rotating machinery applications. In \cite{zhang2020machinery}, a generative adversarial network-based method was utilized to generate additional realistic fake samples, thus solving the class imbalance problem in machinery temporal vibration data. The main drawback of the data-level approach is that during the process of data resampling, important information in the data may be deleted or undesirable noise may be introduced, thus affecting the accuracy of fault diagnosis.

The algorithm-level approach improves the bias against the majority class by making modifications to the algorithm. For example, \citet{he2020spatio} proposed a spatio-temporal multiscale neural network and adopted the focal loss as the loss function to address the data imbalance problem. A data-driven framework for rotating machinery diagnosis was proposed in \cite{xu2020imbalanced} and the imbalanced operating condition was effectively diagnosed via cost-sensitive learning. However, algorithm-level approaches often require a deep understanding of the task in order to select the appropriate algorithm.

NBM methods train the model using only normal data and use the residuals between the measured and model signals as indicators of a possible fault \cite{tautz2017using}. For example, \citet{wei2019wind} designed an NBM method using optimized relevance vector machine regression to predict wind turbine electric pitch system failures using SCADA information. In order to detect wind turbine generator faults, normal behavior models were proposed in \cite{bi2017detection} based on performance curves. Although NBM methods are capable of coping with the data imbalance problem, a limitation of the NBM methods is their inability to diagnose multiple classes of faults.

In this paper, we innovatively propose a supervised contrastive learning method enabled by a hard sample mining framework for wind turbine pitch system fault diagnosis, which successfully solves the hard sample problem of difficulty in distinguishing between multiple health conditions.

\subsection{Motivation}
\begin{figure}[!t]
\centering
\subfloat[]{\includegraphics[width=1.7in]{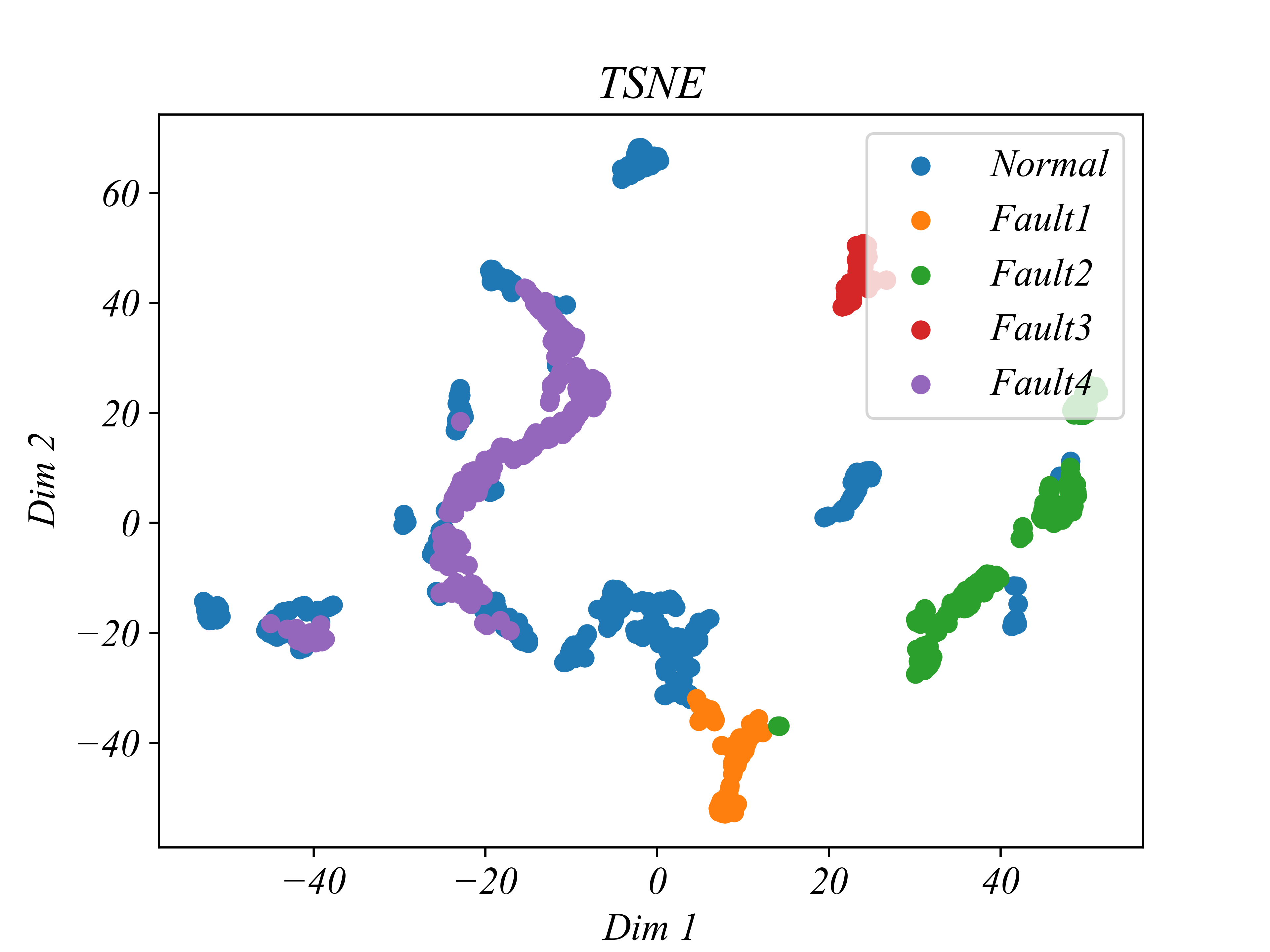}%
\label{motivation_tsne}}
\hfil
\subfloat[]{\includegraphics[width=1.5in]{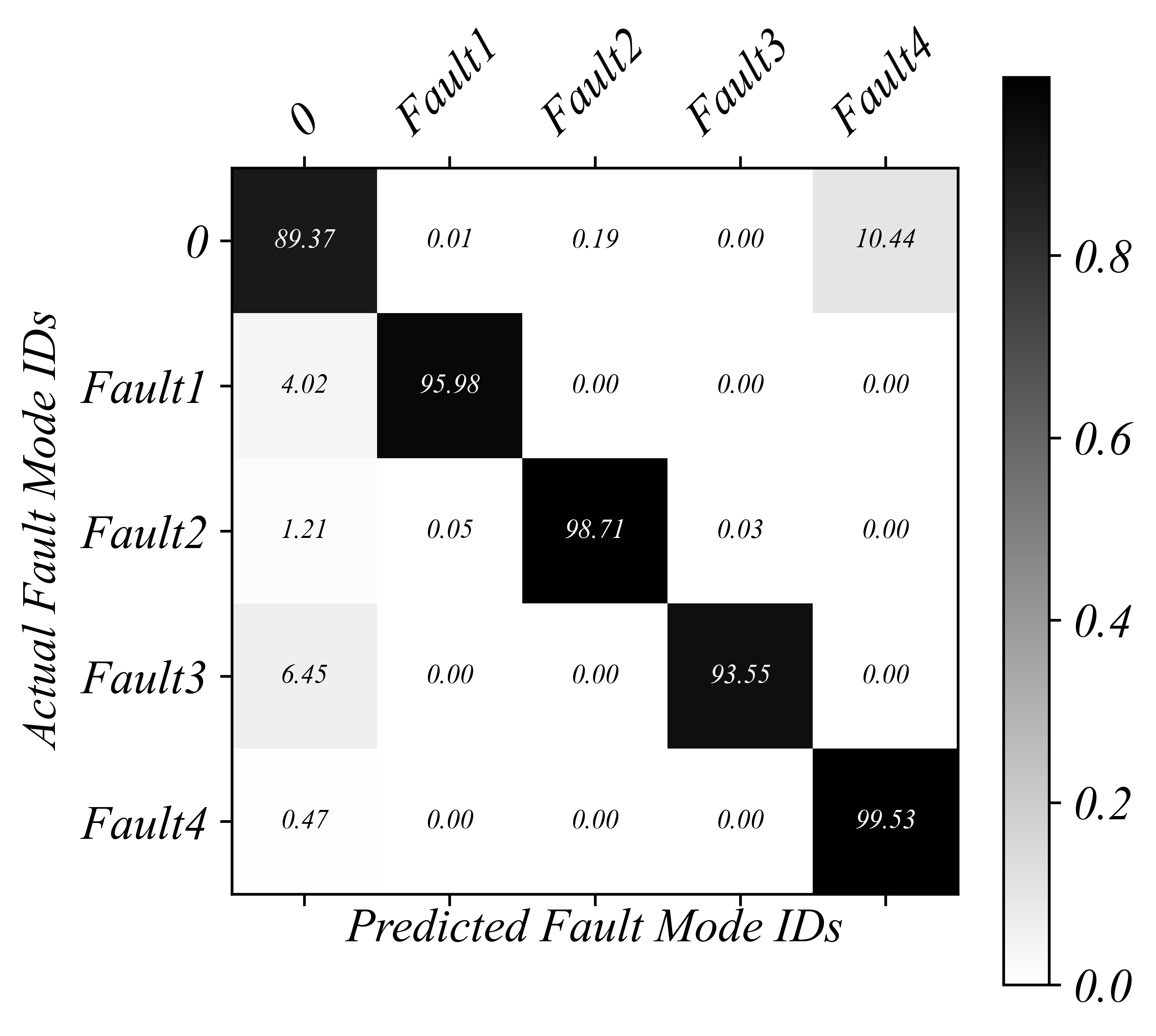}%
\label{motivation_confusion}}
\caption{Fault diagnosis result of wind turbine pitch system by SCL+MLP. (a) TSNE embedding visualization (b) Confusion matrix}
\label{motivationfig}
\end{figure}

The components in the pitch system of a wind turbine gradually wear out during an increasing number of operations, making it challenging to accurately diagnose the health condition of the pitch system components. 
To demonstrate the existence of this phenomenon, we selected a real SCADA dataset of a wind turbine pitch system (This dataset is described in detail in Section \ref{secdatadesc}.) to diagnose its five health conditions by SCL+MLP, the results of which are shown in Fig. \ref{motivationfig}. 
From the figure, it can be seen that there is a serious misclassification problem between the normal class and the four faulty classes. This problem may occur because the health conditions of the pitch system components change gradually over time, and in addition, there is a class imbalance as the different health conditions last for different periods of time. This problem may lead to higher maintenance costs or even damage to the whole wind turbine.

Although data-driven fault diagnosis techniques are widely used for wind turbine pitch systems, to the best of our knowledge, few of them have explored the phenomenon of hard samples between pitch system health conditions. In this paper, we propose a hard sample mining framework based on cosine similarity, which can construct hard sample pairs in a mini-batch to make the training stage more challenging and make the distribution of samples in the feature space more compact and discriminative. Combining the proposed hard sample mining framework with SCL-MLP can well solve the hard sample problem in wind turbine pitch systems. The details of the proposed method will be described in Section \ref{proposedmethod}.

\subsection{Supervised Contrastive Feature Learning}

With its powerful feature extraction capability, contrast learning has been widely used as an efficient feature learning method in natural language processing \cite{gunel2020supervised,inoue2020semi}, computer vision \cite{Wang_2021_CVPR,li2021dual} and other fields. 
Supervised contrastive learning selects an anchor and chooses samples with the same label as the anchor as positives and samples with different labels as negatives. For each anchor, more discriminative representations are learned by pushing the anchor closer to the positives and farther away from the negatives in the embedding space \cite{khosla2020supervised}. The supervised contrastive loss is defined as follows:

\begin{equation}
    \mathcal{L}^\text{sup} = \sum_{i \in I} \frac{-1}{|P(i)|} \sum_{p \in P(i)} log \frac{e^{sim({z_i}, {z_p}) / \tau}}{\sum_{a \in A(i)}
    e^{sim({z_i}, {z_a})/\tau} }
    \label{losssup}
\end{equation}
where $I \in D^l$ is a mini-batch of the labeled data, and $\tau \in \mathcal{R}^{+}$ is a scalar temperature parameter. $P(i)\equiv\left\{p \in A(i): \tilde{\boldsymbol{y}}_p=\tilde{\boldsymbol{y}}_i\right\}$ is the set of indices of all positives of the anchor and $|P(i)|$ is its cardinality. 
Supervised contrastive learning leverages label information effectively, enabling models trained with supervised contrastive loss to acquire compact representations. Consequently, these models can capture more discriminative features. 

\subsection{Hard Sample Mining}

\begin{figure}[!t]
\centering
\includegraphics[width=0.95\linewidth]{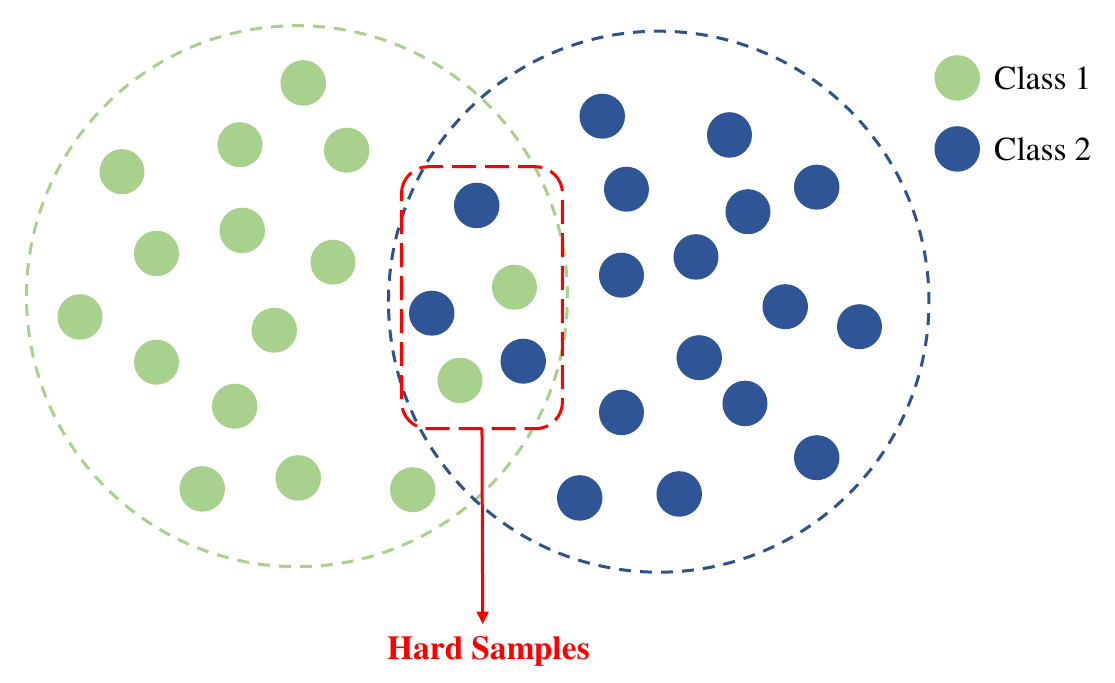}
\caption{The schematic diagram of hard samples.}
\label{hardsample}
\end{figure}

The boundaries of samples of different classes in the feature space may be close to each other, so samples near the boundaries will be difficult to classify, and such samples are hard samples. A schematic diagram of hard samples is shown in Fig. \ref{hardsample}, where the two classes are represented by green and blue dots, respectively, and the data points in the region framed by the red dotted line are the hard samples. Moreover, if a sample is selected as the query sample, the hard samples of the query sample include hard positive samples with the same label but far away from it and hard negative samples with different labels but close to it \cite{smirnov2018hard,chen2020hard}. Making full use of hard samples in the training phase of the model not only improves the learning efficiency but also enables the model to have better classification performance. 

 Hard sample mining has been widely used in computer vision and other fields \cite{shrivastava2016training,sun2019mvp,zhu2019distance}. 
\citet{Suh_2019_CVPR} proposed a stochastic hard negative mining method that adopts class signatures to keep track of feature embedding online. The method improves image retrieval accuracy substantially. In \cite{Sun_2019_ICCV}, a weighted complete bipartite graph-based maximum-value perfect matching for mining the hard samples was proposed to relieve adverse optimization and sample imbalance problems. \citet{xue2019hard} applied a hard sample mining method based on an enhanced deep multiple instance learning to find the hard samples from unlabeled training data. 
However, few hard sample mining methods have been employed in the field of fault diagnosis for wind turbines, and the method proposed in this study fills the gap in the field.

\begin{figure*}[htb]
\centering
\includegraphics[width=0.95\linewidth]{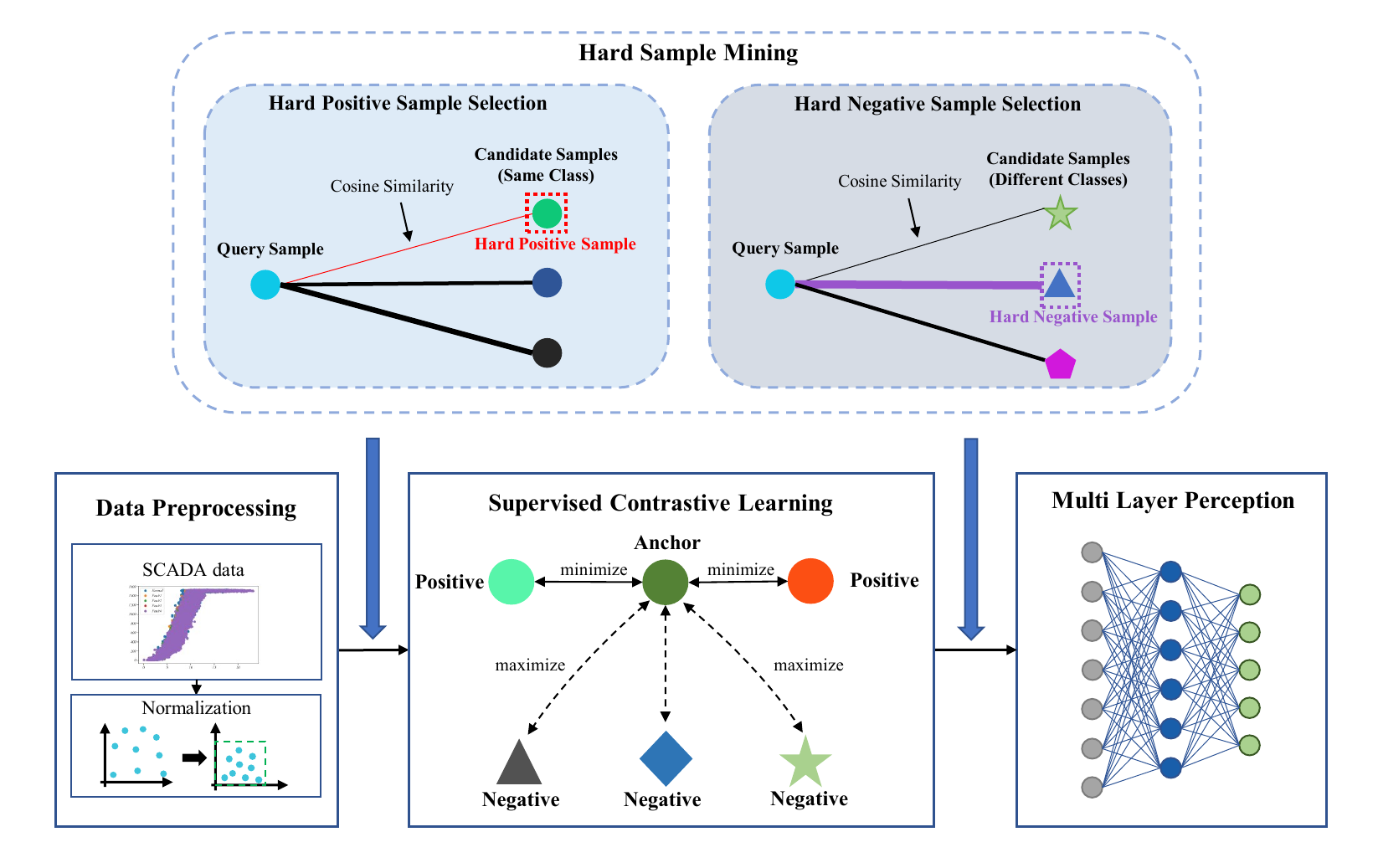}
\caption{Overview of proposed HSMSCL}
\label{overview}
\end{figure*}

\begin{algorithm}[htb]
\caption{Hard Sample Mining Enabled Supervised Contrastive Learning}

\label{hsmsclalgorithm}
\SetKwProg{Function}{Function}{}{}

\KwIn{labeled train set $(x,y) \in D$, learning rate $\eta$, batch\_size $bs$, epochs $n$, temperature $\tau$, model $f$ and $g$ with parameters $\theta$ and $\theta_g$, number of HSM stages $num\_stages$}
\KwOut{trained model with parameters $\theta^{*}$}

Net, $\theta \leftarrow \text{initialize}(Net)$\;
$x \leftarrow \text{Normalize}(x)$\;

\Function{\text{HSM(bs, D, mis\_samples)}}
{
    hard\_batches $\leftarrow$ []\;
    \For{$b = 1$ \KwTo $length(D) // bs$}
    {
        batch $\leftarrow$ []\;
        batch $\leftarrow$ Initialization(batch, $D_i$, mis\_samples)\;
        left $\leftarrow$ bs - length(batch)\;
        \While{$left > 0$}
        {
            append(batch, random\_sample($D$))\;
            s $\leftarrow$ random\_sample(batch)\;
            append(batch, hard\_positive($s$))\;
            append(batch, hard\_negative($s$))\;
        }
        append(hard\_batches, batch)\;
    }
    \Return hard\_batches\;
}
\underline{\textbf{Stage 1:} \textit{SCL}}

\For{$stage = 1$ \KwTo $num\_stages$}
{
    hard\_batches $\leftarrow$ HSM(bs, D, [])\;
    \For{$epoch = 1$ \KwTo $n$}
    {
        \For{$(x_b,y_b) \in hard\_batches$}
        {
            $out_{b} \leftarrow f(\theta, x_{b})$\;
            $\mathcal{L} \leftarrow$ loss\_SCL$(y_b, out_b, \tau)$\;
            $\theta \leftarrow$ back\_propagation$(\mathcal{L}, \theta, \eta)$\;
        }
    }
}
\underline{\textbf{Stage 2:} \textit{MLP}}

mis\_samples $\leftarrow$ []\;

\For{$stage = 1$ \KwTo $num\_stages$}
{
    hard\_batches $\leftarrow$ HSM(bs, D, mis\_samples)\;
    \For{$epoch = 1$ \KwTo $n$}
    {
        \For{$(x_b,y_b) \in hard\_batches$}
        {
            $out_{b} \leftarrow g(\theta_g, x_{b})$\;
            $\mathcal{L} \leftarrow$ loss\_MLP$(y_b, out_b, \tau)$\;
            $\theta_g \leftarrow$ back\_propagation$(\mathcal{L}, \theta_g, \eta)$\;
        }
    }
    mis\_samples $\leftarrow$ misclassification(g, D)\;
}
\end{algorithm}

\section{Proposed method}
\label{proposedmethod}
\subsection{Overview}

The structure of the proposed method is shown in Fig.\ref{overview}. The proposed method consists of five parts: data preprocessing, data augmentation, hard sample mining, supervised contrastive learning, and multi-layer perceptron.

\begin{enumerate}
\item Data preprocessing

SCADA data is collected by hundreds of various types of sensors inside the wind turbine, therefore, the raw data needs to be normalized in order to eliminate the effects of different variable sizes. In this paper, we use Minimum-Maximum Normalization, which individually scales each data feature to [0,1] by a linear transformation of the original data. 
The formula of Minimum-Maximum Normalization is shown as follows:
\begin{equation}
\begin{aligned}
    x_i^{\text {norm }}=\frac{x_i-x_i^{\min }}{x_i^{\max }-x_i^{\min }}
\end{aligned}
\label{normalization}
\end{equation}
where $x_i$ is the $i$th feature of the input sample and $x_i^{\max}$ and $x_i^{\min}$ are the maximum and minimum values found in $x_i$, respectively.





\item Hard sample mining(HSM)

In this paper, the proposed HSM method based on cosine similarity is used to select the hard samples. Specifically, when constructing a mini-batch, the proposed HSM method first randomly selects a sample from each class as initialization. Subsequently, either a random sample from the dataset is selected or a random query sample is chosen from the current mini-batch and its hard samples are determined to extend this mini-batch. In the latter case, the sample of a different class with the highest cosine similarity to the selected query sample is selected as its hard negative sample, while the sample of the same class with the lowest cosine similarity to the query sample is selected as a hard positive sample. We keep repeating the above process until the size of the mini-batch reaches the batch size. It is worth noting that the proposed HSM method is applied to construct the mini-batches of both the SCL and MLP stages. In the MLP stage, specifically, the misclassified samples for the current stage are used by HSM to initialize the mini-batches in the next stage. In this case, each time at initialization, a sample is randomly taken from misclassified samples if it is not empty, or from the dataset otherwise. 


\item Supervised Contrastive Learning(SCL)

With the powerful representation learning capability of SCL, the proposed method is capable of learning the rich representations in the data. In addition, the proposed method combines HSM with SCL by utilizing HSM to construct more challenging mini-batches, thus allowing SCL to learn more discriminative representations.


\item Multi-layer perceptron(MLP)

The representations learned by SCL will be constructed by HSM into more challenging mini-batches for training MLP to classify the input data.

\end{enumerate}

\subsection{Hard sample mining enabled supervised contrastive learning(HSMSCL)}


The key to the proposed HSMSCL is to combine HSM as a unified framework with SCL and MLP, allowing SCL to learn more compact representations and train better MLP classifiers for more accurate diagnosis of hard-to-classify faults in wind turbine pitch systems. As shown in detail in Algorithm\ref{hsmsclalgorithm}, the two stages, SCL and MLP, each invoke HSM as a framework to construct more challenging mini-batches. 

By introducing hard samples through HSM, SCL enhances the learning of the model for sample similarity and difference, which in turn improves the model's performance in the fault diagnosis task. Through multiple rounds of training, the model gradually learns more discriminative features, which provides a better basis for feature representation in the subsequent MLP stage.

The main purpose of the MLP stage is to mine hard samples by using HSM and use these hard samples for training the MLP model. Hard samples introduce challenges in the training process and can motivate the model to better adapt to complex situations, thus improving the overall performance. Repeating the process of multiple rounds of training and updating the list of misclassified samples allows the MLP model to progressively optimize itself and enhance its accuracy and robustness to complex fault diagnosis tasks.

Overall, the proposed HSMSCL improves the performance of the fault diagnosis model for hard samples by utilizing the HSM framework based on the two-step approach of SCL and MLP.

\begin{table}[htb]
\centering
\caption{26 parameters of SCADA data.}
\label{variable}
\begin{tabular}{l|l}
\toprule
No. & Parameter                            \\ \hline
1   & Wind speed                           \\
2   & Generator speed                      \\
3   & Active power                         \\
4   & Wind direction                       \\
5   & Average wind direction within 25s    \\
6   & Yaw position                         \\
7   & Yaw speed                            \\
8   & Angle of pitch 1                     \\
9   & Angle of pitch 2                     \\
10  & Angle of pitch 3                     \\
11  & Speed of pitch 1                     \\
12  & Speed of pitch 2                     \\
13  & Speed of pitch 3                     \\
14  & Temperature of pitch motor 1         \\
15  & Temperature of pitch motor 2         \\
16  & Temperature of pitch motor 3         \\
17  & Horizontal acceleration              \\
18  & Vertical acceleration                \\
19  & Environment temperature              \\
20  & Internal temperature of nacelle      \\
21  & Switching temperature of pitch 1     \\
22  & Switching temperature of pitch 2     \\
23  & Switching temperature of pitch 3     \\
24  & DC current of pitch 1 switch charger \\
25  & DC current of pitch 2 switch charger \\
26  & DC current of pitch 3 switch charger \\ 
\bottomrule
\end{tabular}
\end{table}

\section{Experiment}

\begin{figure}[htb]
\centering
\subfloat[]{\includegraphics[width=2.5in]{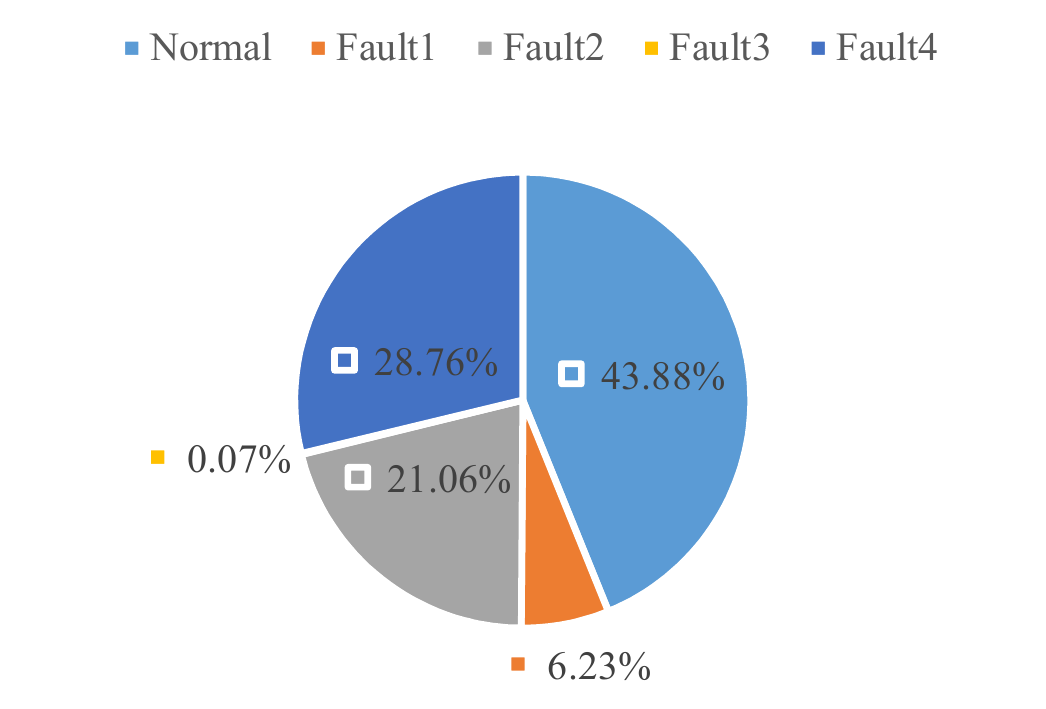}%
\label{datadistribution23}}
\hfil
\subfloat[]{\includegraphics[width=2.5in]{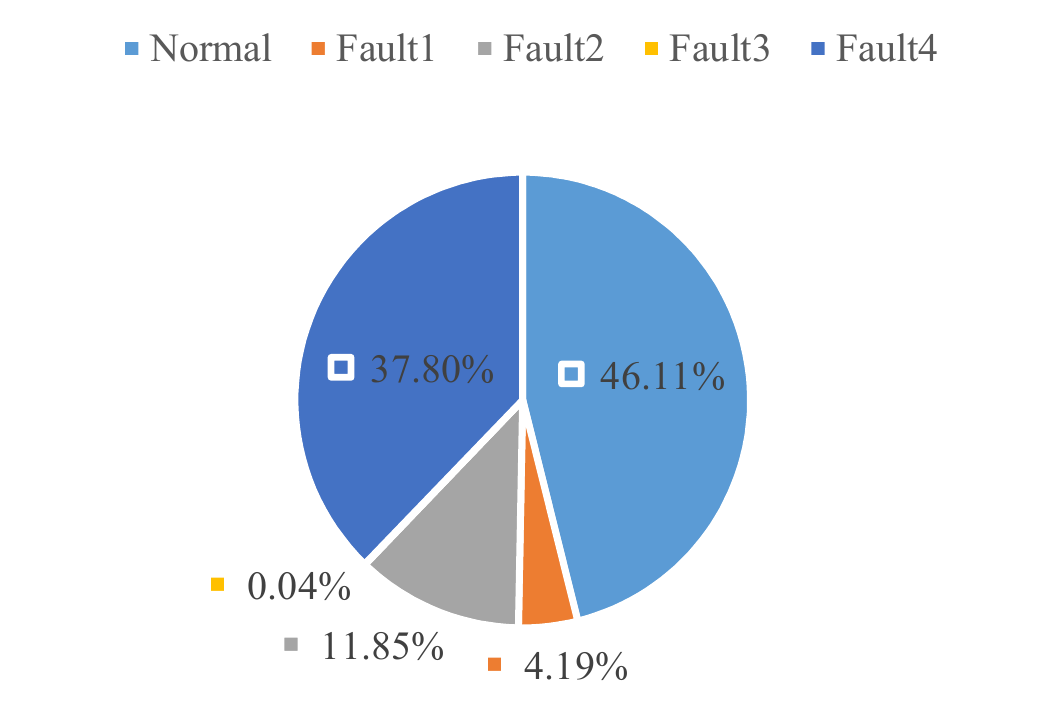}%
\label{datadistribution29}}
\caption{The data distribution of the two datasets. (a) WT 23 (b) WT 29}
\label{datadistribution}
\end{figure}

\begin{figure}[htb]
\centering
\subfloat[]{\includegraphics[width=2.5in]{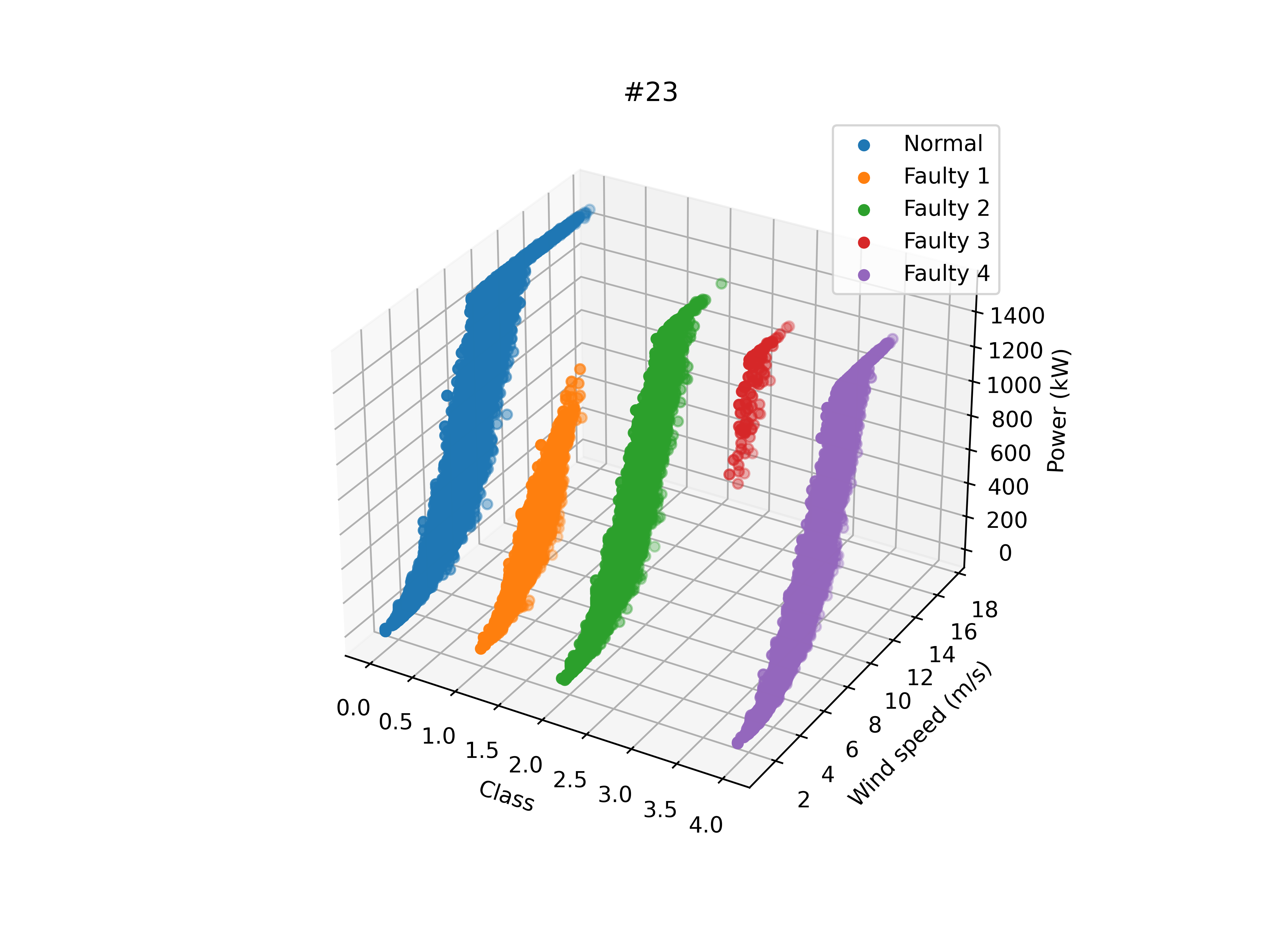}%
\label{power23}}
\hfil
\subfloat[]{\includegraphics[width=2.5in]{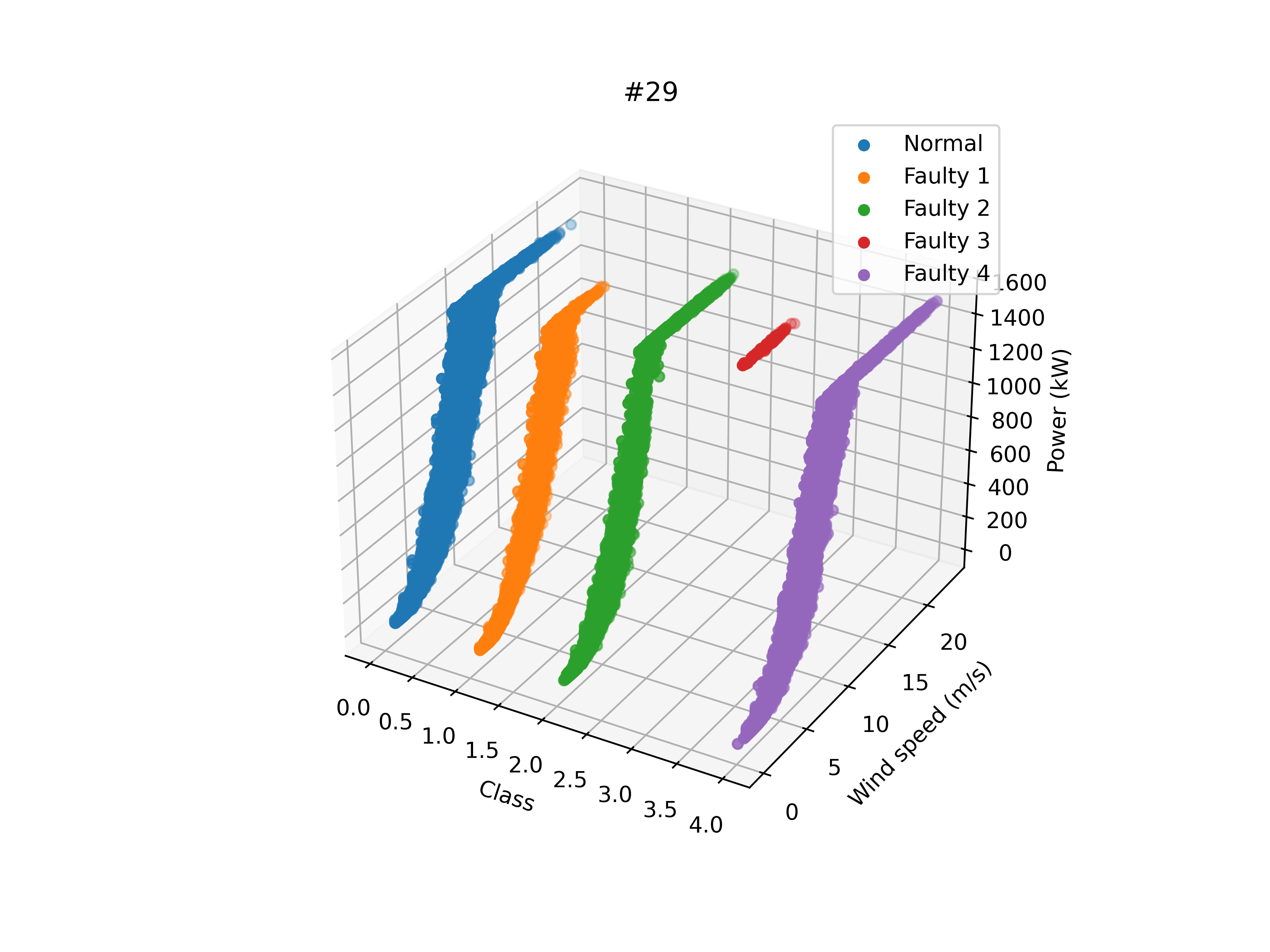}%
\label{power29}}
\caption{The output power VS wind speed for WT 23 and WT 29. (a) WT 23 (b) WT 29}
\label{power}
\end{figure}

\subsection{Experimental Setup}
\subsubsection{Data Description}
\label{secdatadesc}
To validate the effectiveness of the proposed method, two pitch system cog belt fracture datasets collected from two real-world wind turbines numbered 23 and 29 are used \cite{industrybigdata} (in the remainder of this paper, they are referred to as WT 23 and WT 29). A cog belt is a transmission device with a toothed structure on the working surface and is a key component in the electric pitch system of a wind turbine. During wind turbine operation, the cog belt may gradually break due to stress fatigue, leading to the failure of the wind turbine pitch system. There are five health conditions in the datasets: normal, slightly worn, low risk of fracture, high risk of fracture, and complete fracture. If the cog belt completely fractures, it can significantly affect the wind turbine's power production and increase maintenance costs. Therefore, it is crucial to accurately diagnose the health condition of the cog belt and replace it before it fractures completely to ensure the normal operation of the wind turbine. 


The original sensor data contains hundreds of variables recorded by a large number of sensors of the SCADA system inside the wind turbine. After selection by experts in the field, the remaining irrelevant parameters were removed and 26 parameters related to the cog belt fracture of the wind turbine were retained. The details of the parameters are shown in Table \ref{variable}. The total number of data for WT 23 and WT 29 is 13,6709 and 215,012, respectively. Fig.\ref{datadistribution} shows the extreme imbalance in the duration of the different health conditions in the datasets. Fig.\ref{power} shows the output power VS wind speed for WT 23 and WT 29. It can be seen that the cog belt has different effects on the power generation efficiency of wind turbine when it is in different health conditions. During the experiment, we divided 70\% of the data into the training set and 30\% into the test set, where 20\%  of the training set is further divided as the validation set.


\subsubsection{Evaluation Metrics}
In this paper, In this paper, \text{Accuracy} and \text{macro G-mean} \cite{barandela2003strategies,ri2020g} are used as evaluation metrics. \text{G-mean} is a good indicator of the classifier's performance in imbalanced classification problems \cite{kubat1997addressing}. Due to the extreme class imbalance in the dataset used in this paper, the macro average of \text{G-mean} was used in order to fully represent the effectiveness of different methods in classifying each health condition. The evaluation metrics are defined as follows.

\begin{equation}
\text{Accuracy} = \frac{\mathrm{TP}+\mathrm{TN}}{\mathrm{TP+FP+FN+TN}}
\label{acc}
\end{equation}

\begin{equation}
\text{G-mean} = \sqrt{ \frac{\mathrm{TP}}{\mathrm{TP+FN}} \times \frac{\mathrm{TN}}{\mathrm{TN+FP}}}
\label{gmean}
\end{equation}

\begin{equation}
\operatorname{macro} \text{ G-mean}=\frac{1}{n} \sum_{i=1}^n \mathrm{G-mean_i}
\label{macrogmean}
\end{equation}
where \text{TP}, \text{TN}, \text{FP}, and \text{FN} refer to true positives, true negatives, false positives, and false negatives, respectively.

\subsubsection{Baseline}

To verify the effectiveness of the proposed method, we selected 5 baselines, namely SCL-MLP (with the HSM framework removed from the proposed method), ResNet101, BiLSTM, and CNN. In order to solve the problem of extreme class imbalance in the datasets, Focal Loss (FL) \cite{lin2017focal} was used in ResNet101, BiLSTM, and CNN. A brief description of the baselines is given below.

\begin{enumerate}
\item SCL-MLP is the method after removing the HSM framework from the proposed method.
\item ResNet101 is a residual network with a depth of 101 layers.
\item BiLSTM is a recurrent neural network consisting of a forward LSTM and a backward LSTM.
\item CNN used in this paper is a deep neural network consisting of 4 convolutional layers and 3 fully connected layers.
\item MLP used in this paper consists of three fully connected layers.
\end{enumerate}

\begin{figure*}[htbp]
\centering
\subfloat[]{\includegraphics[width=2in]{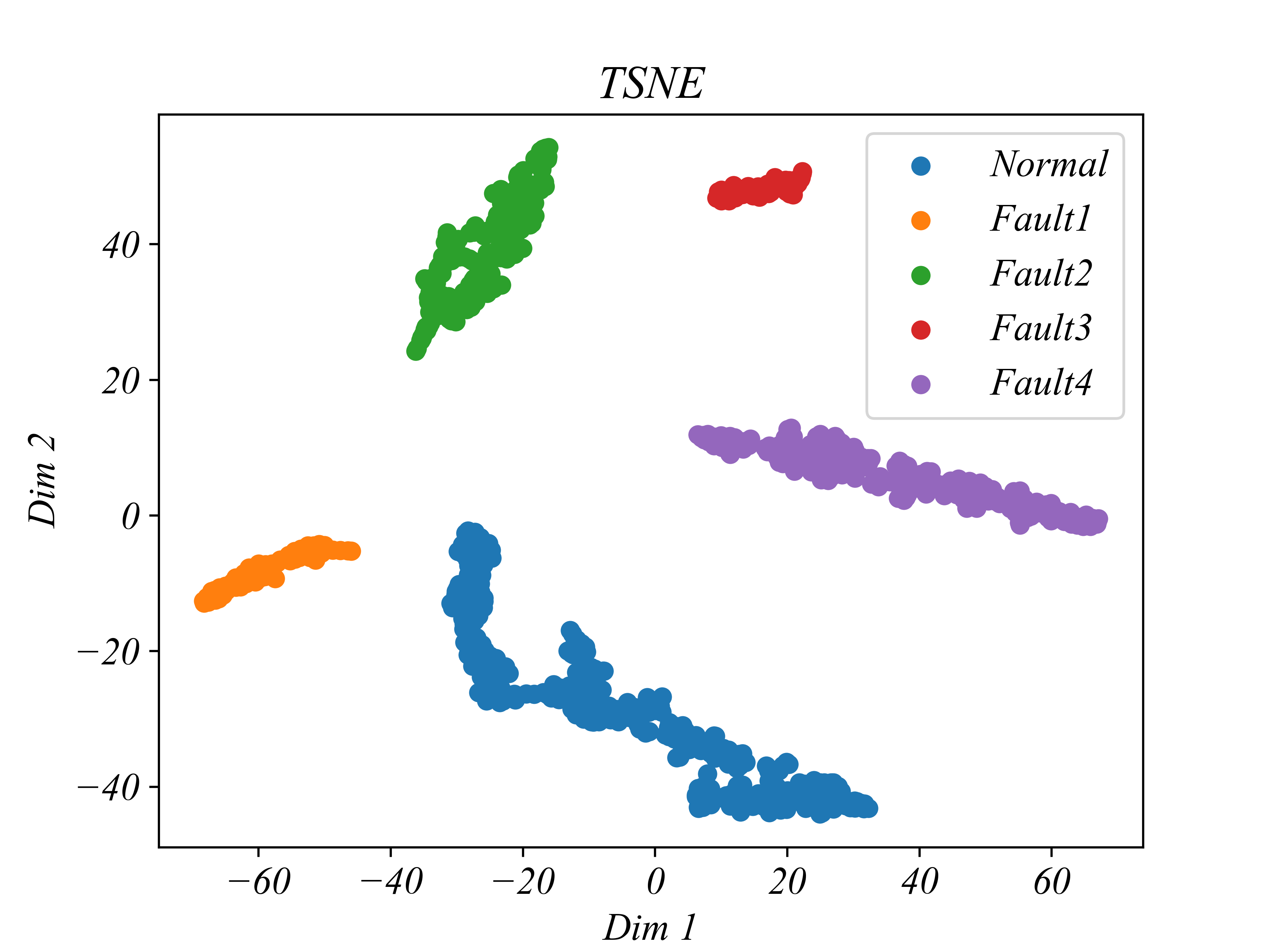}%
\label{TSNE23HSMSCL}}
\hfil
\subfloat[]{\includegraphics[width=2in]{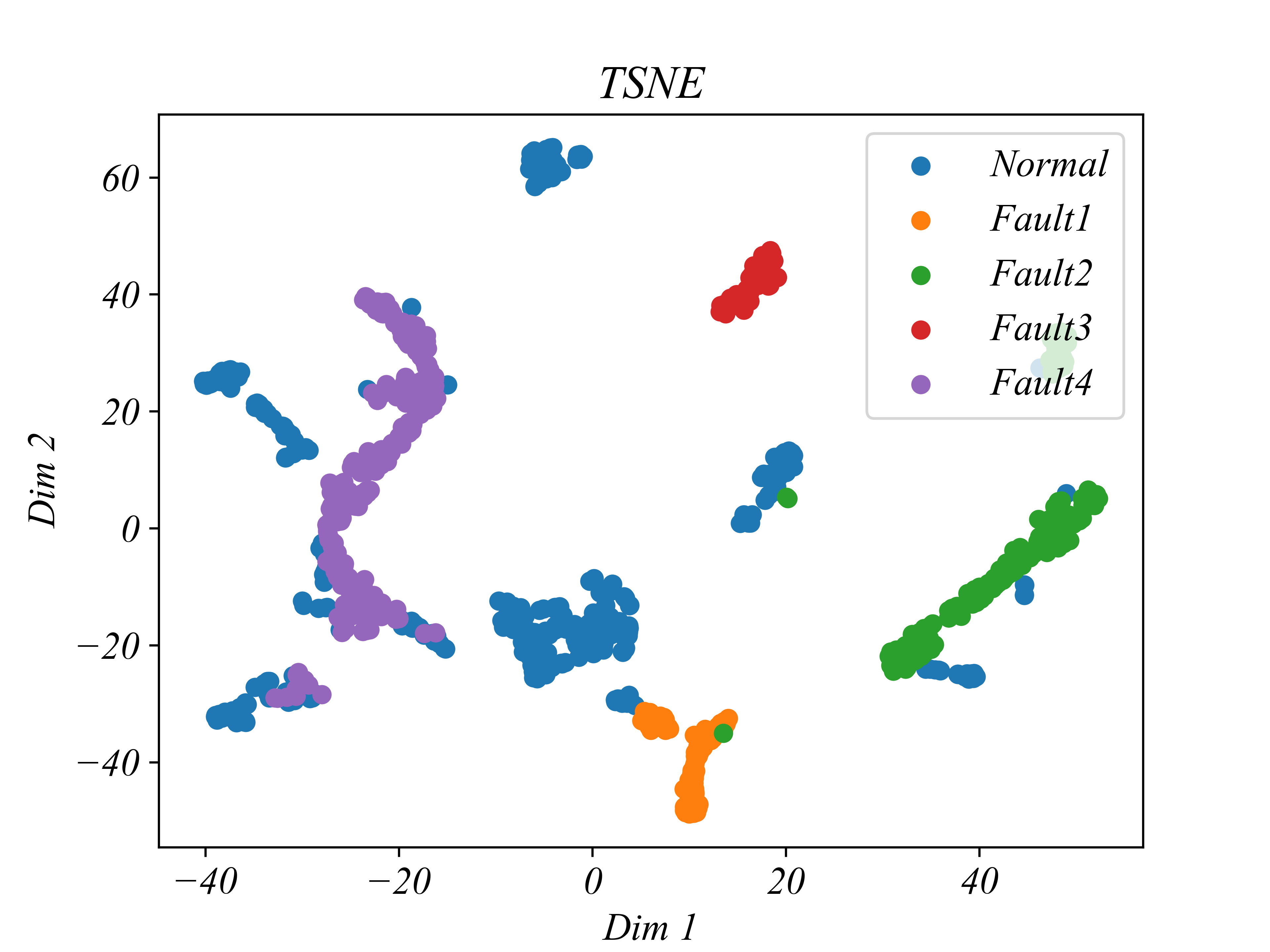}%
\label{TSNE23SCL}}
\hfil
\subfloat[]{\includegraphics[width=2in]{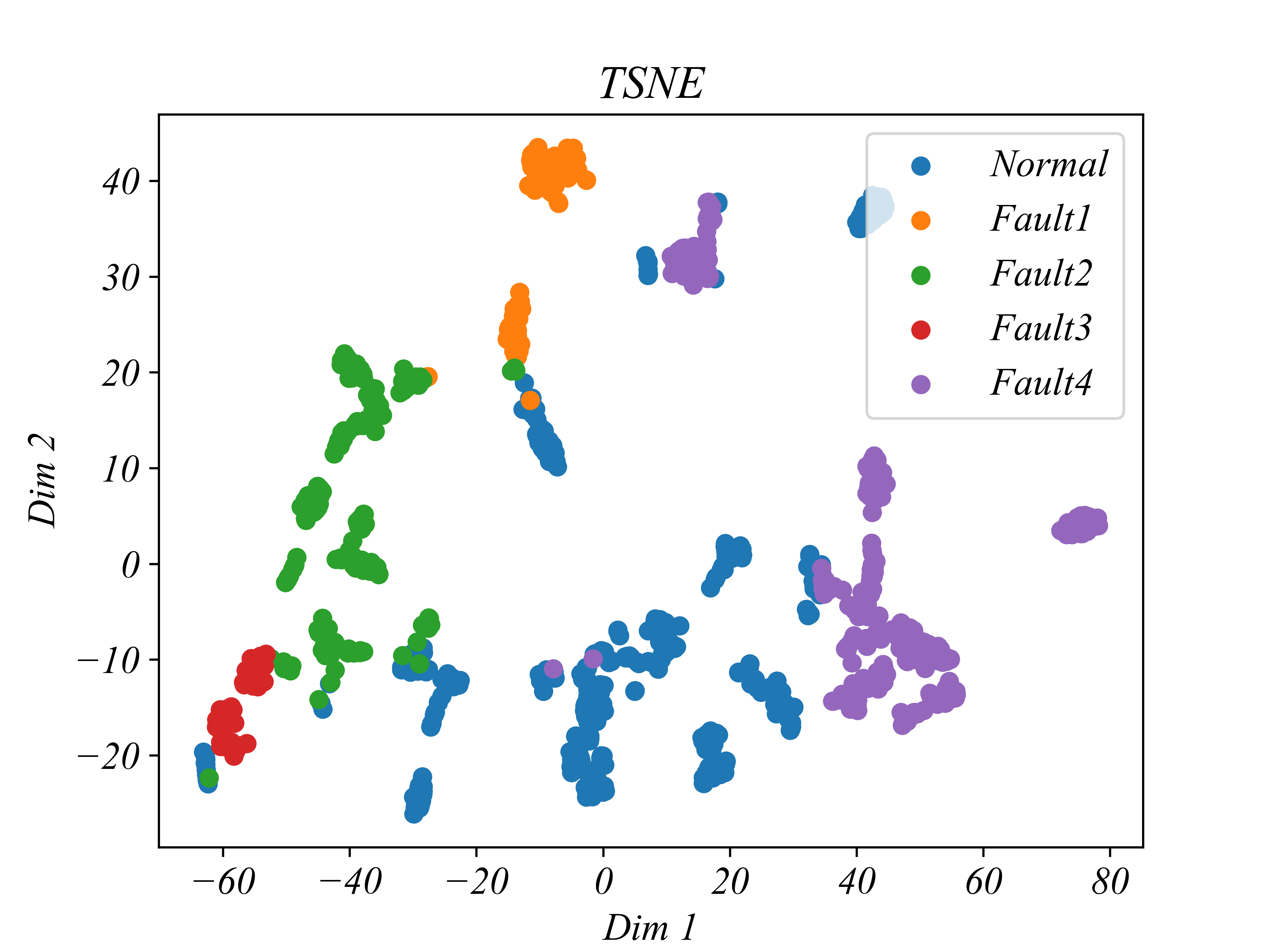}%
\label{TSNE23MLP}}
\hfil
\\
\subfloat[]{\includegraphics[width=2in]{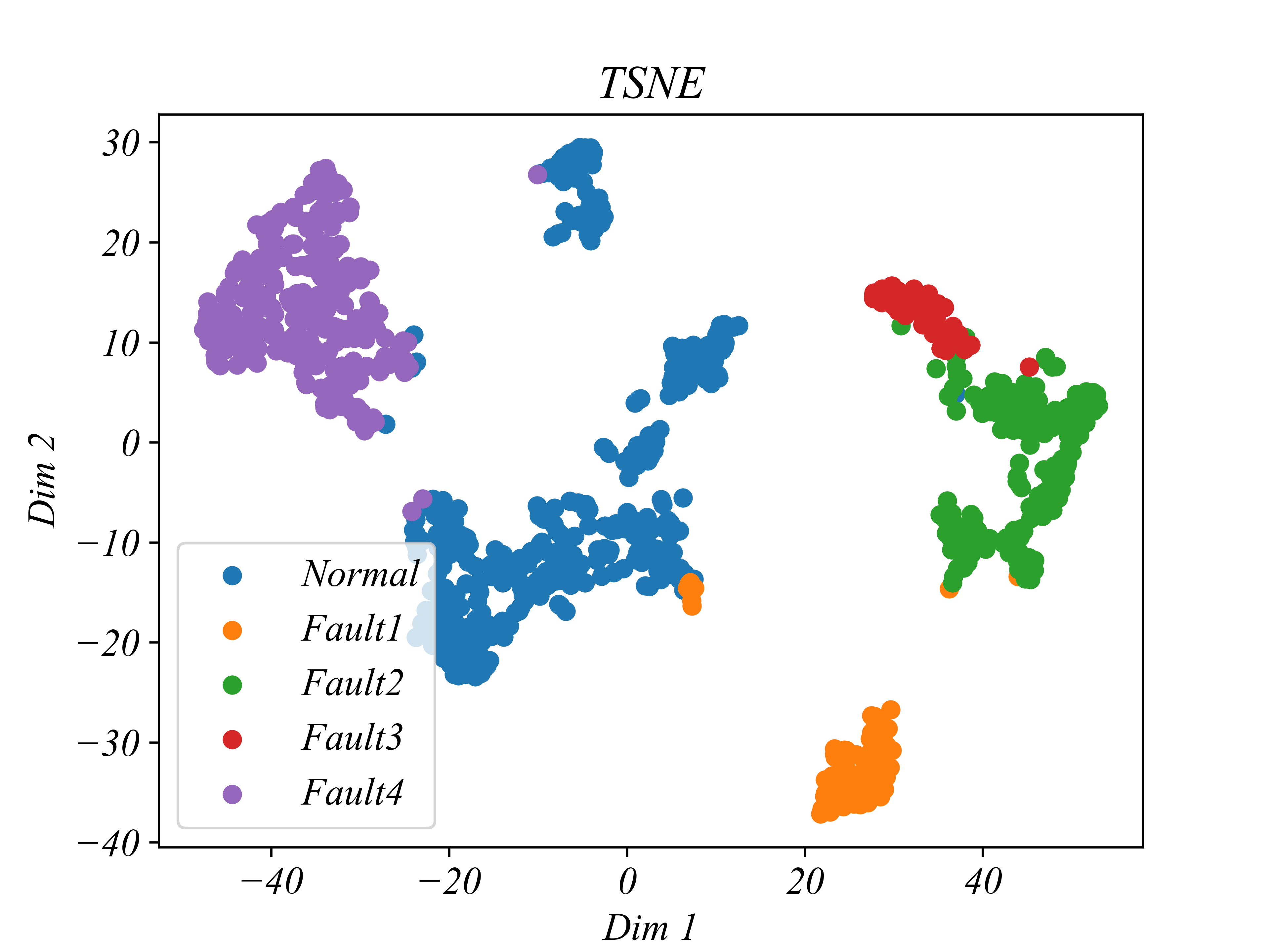}%
\label{TSNE23Resnet}}
\hfil
\subfloat[]{\includegraphics[width=2in]{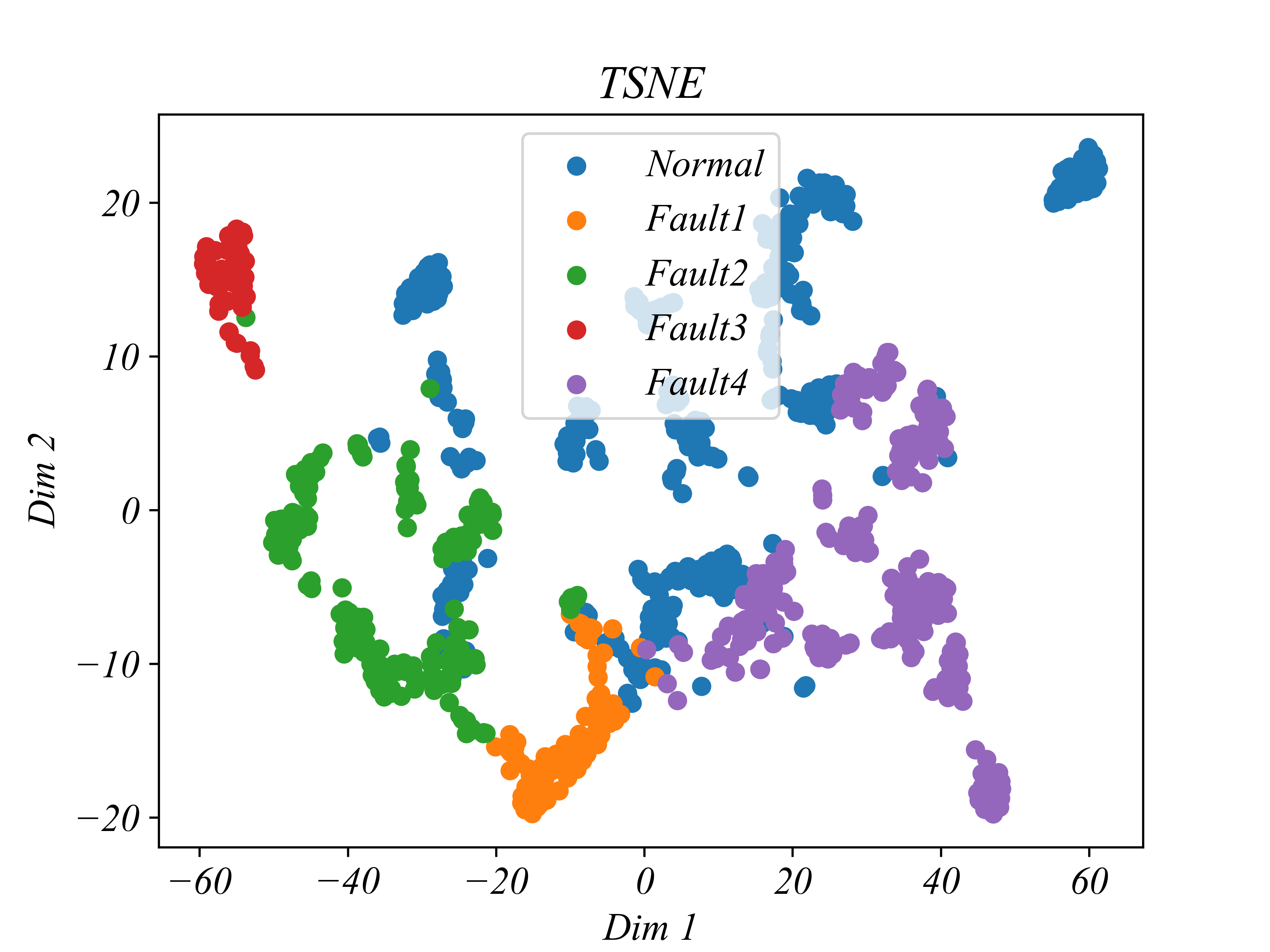}%
\label{TSNE23LSTM}}
\hfil
\subfloat[]{\includegraphics[width=2in]{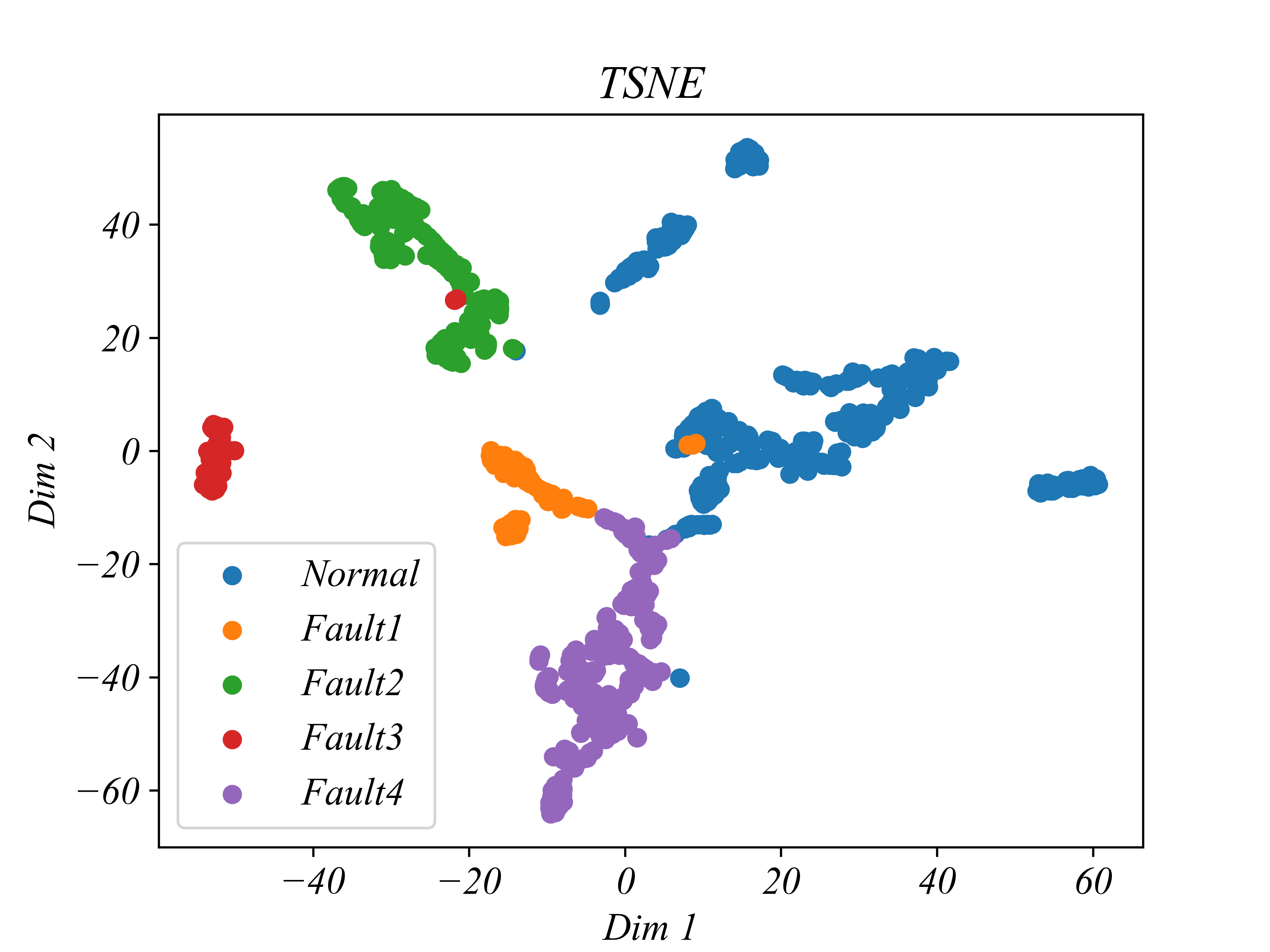}%
\label{TSNE23CNN}}
\caption{T-SNE embedding for WT 23 (a) Proposed method (b) SCL+MLP (c) MLP+FL (d) ResNet101+FL (e) BiLSTM+FL (f) CNN+FL}
\label{TSNE23}
\end{figure*}

\begin{figure*}[htbp]
\centering
\subfloat[]{\includegraphics[width=2in]{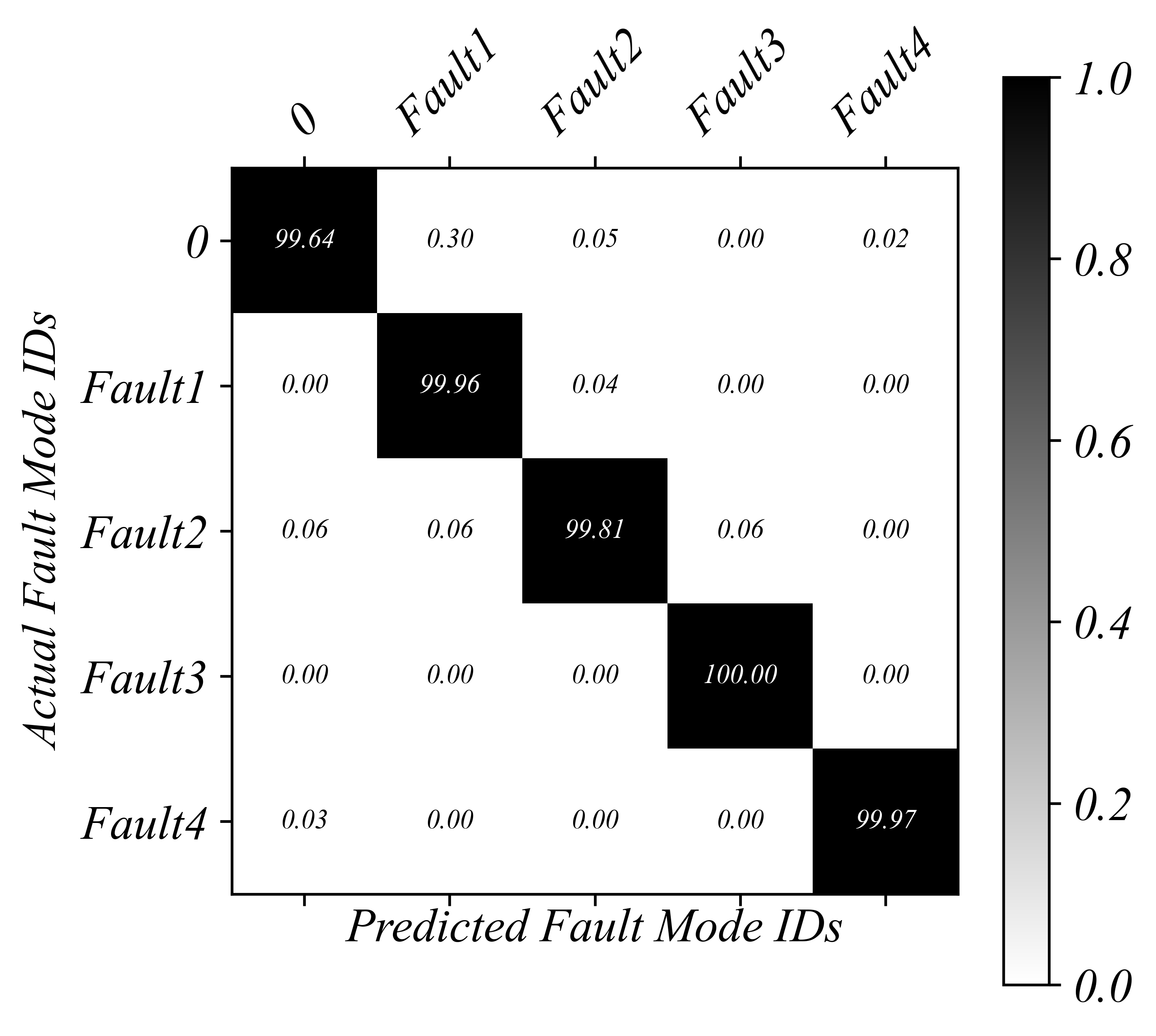}%
\label{confusion23HSMSCL}}
\hfil
\subfloat[]{\includegraphics[width=2in]{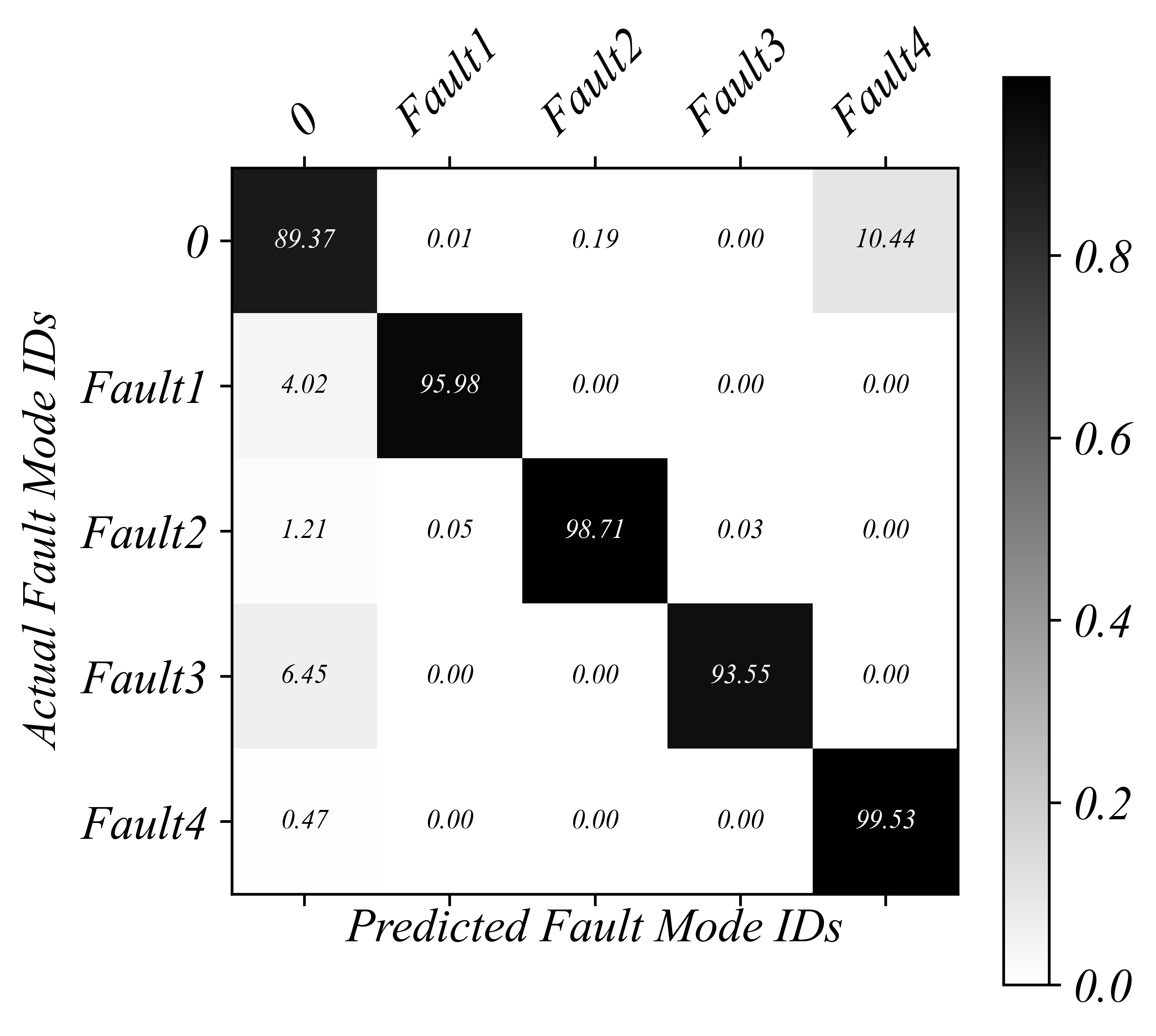}%
\label{confusion23SCL}}
\hfil
\subfloat[]{\includegraphics[width=2in]{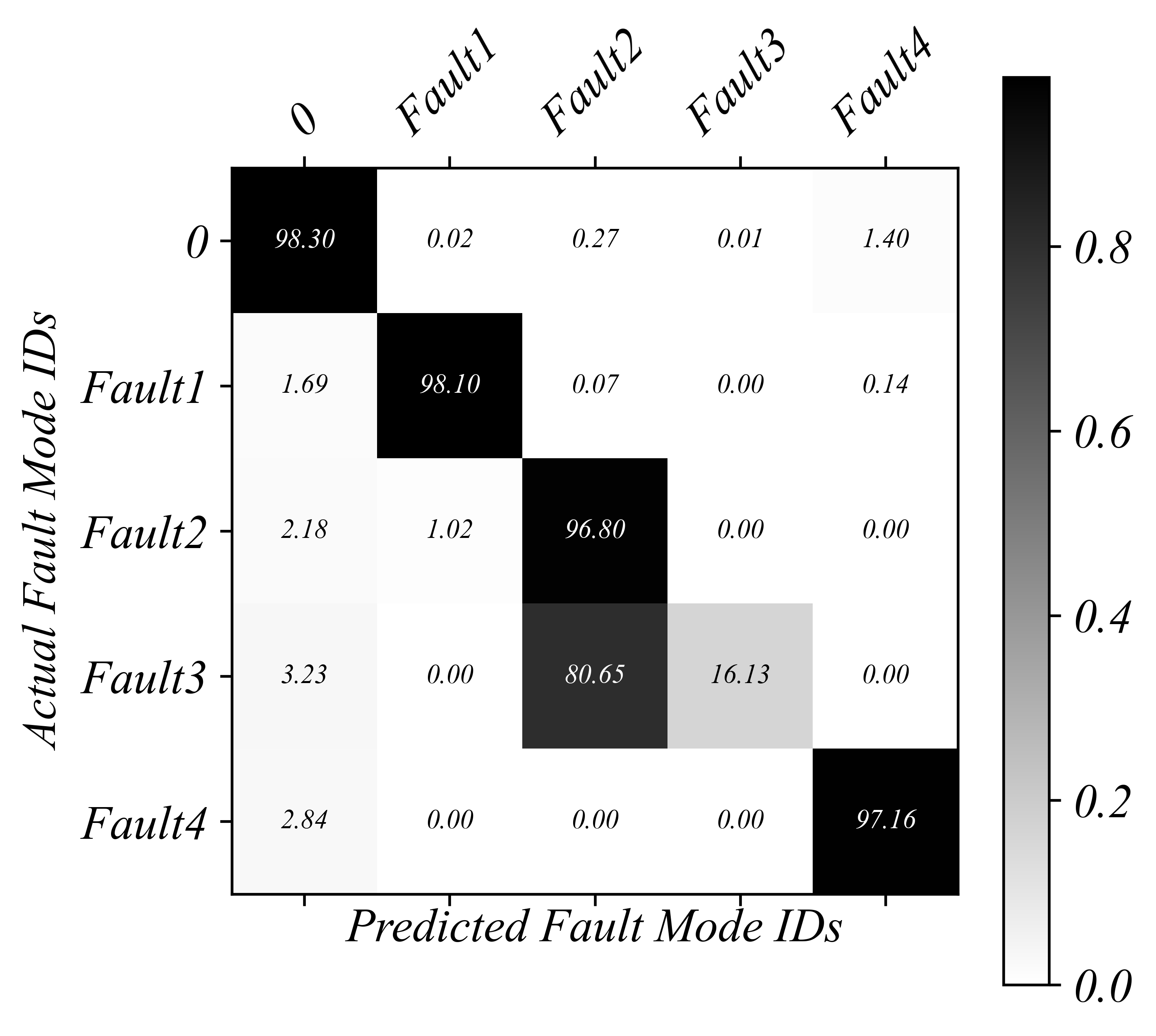}%
\label{confusion23MLP}}
\hfil
\\
\subfloat[]{\includegraphics[width=2in]{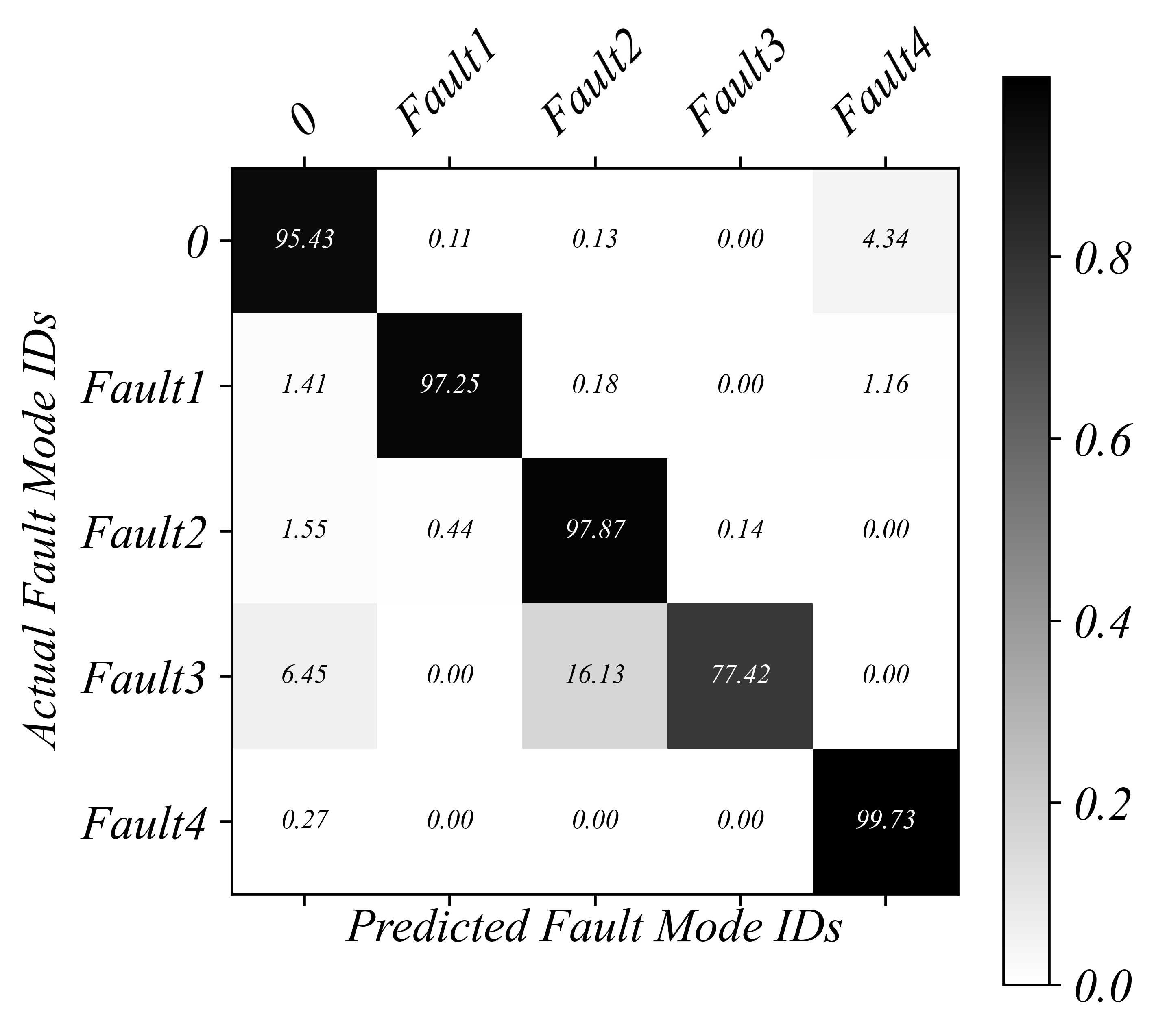}%
\label{confusion23Resnet}}
\hfil
\subfloat[]{\includegraphics[width=2in]{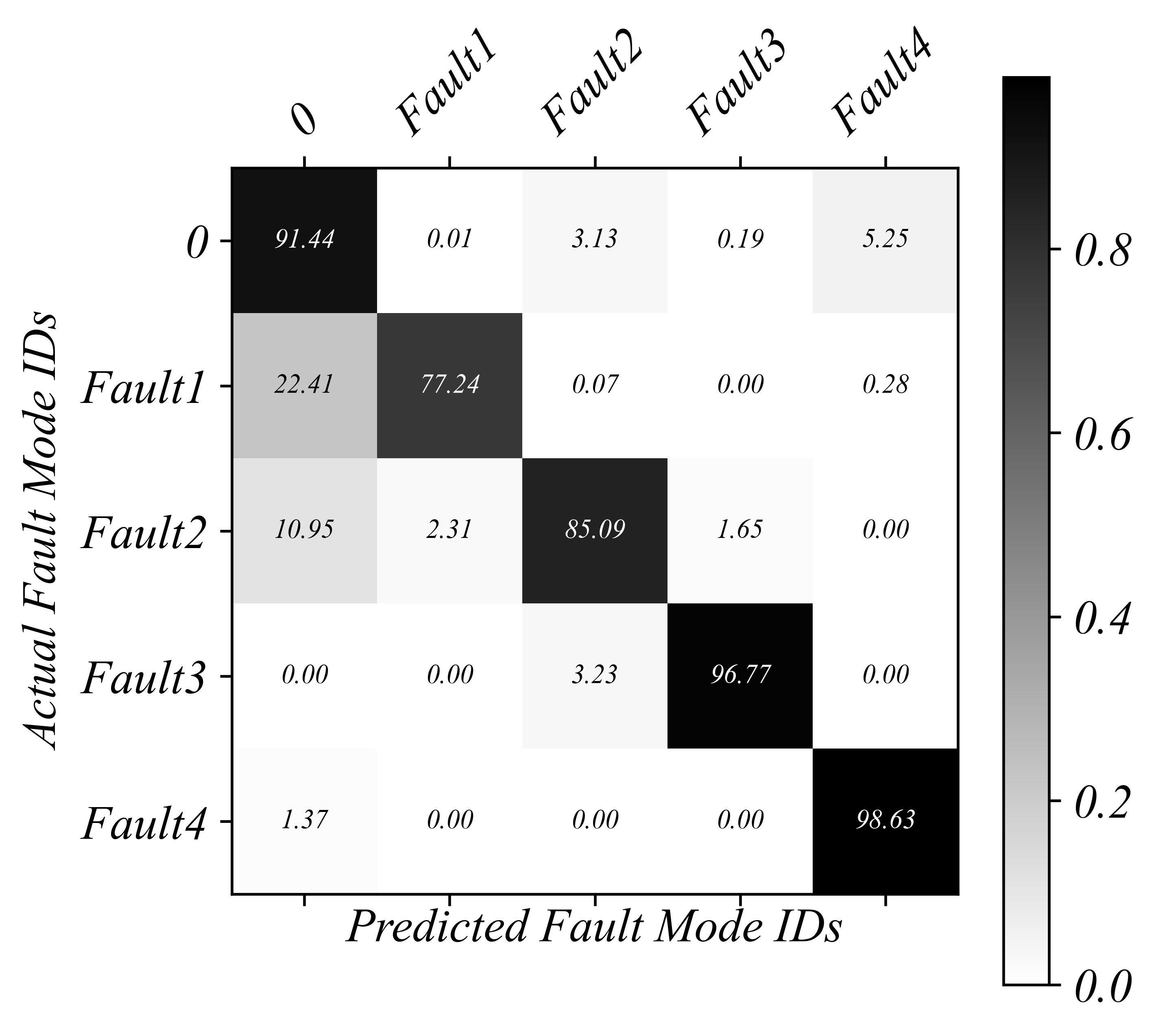}%
\label{confusion23LSTM}}
\hfil
\subfloat[]{\includegraphics[width=2in]{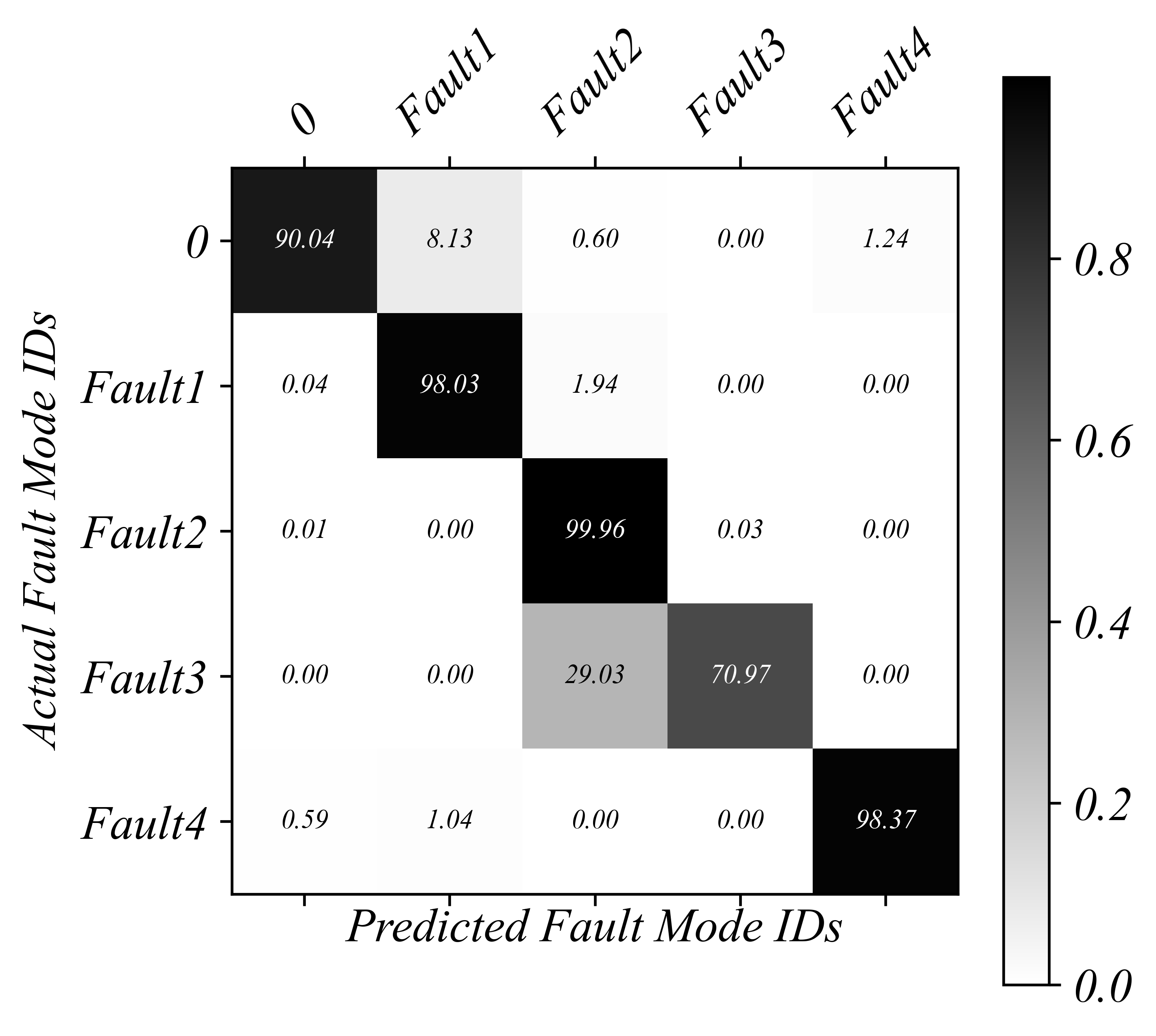}%
\label{confusion23CNN}}
\caption{Confusion matrix for WT 23 (a) Proposed method (b) SCL+MLP (c) MLP+FL (d) ResNet101+FL (e) BiLSTM+FL (f) CNN+FL}
\label{confusion23}
\end{figure*}

\subsection{HSMSCL}

\subsubsection{Experiment for WT 23}
 We compared the proposed HSMSCL and the selected 5 baselines on the dataset of WT 23. We repeated the experiment ten times with different random seeds and the reported results are based on the average. The results are shown in Table \ref{resultWT23}. The corresponding t-SNE embeddings and confusion matrix are shown in Fig.\ref{TSNE23} and Fig.\ref{confusion23}. Compared with the baselines, the $Accuracy$ of the proposed method is increased by 5.41\%, 0.30\%, 0.03\%, 1.17\%, 0.15\% and 0.42\%, and the \text{macro G-mean} of the proposed method is increased by 3.07\%, 7.02\%, 3.29\%, 6.70\%, 9.43\% and 2.27\%. The results in Table \ref{resultWT23} show that the proposed HSMSCL performs significantly better than baselines, especially in the value of \text{macro G-mean}. Fig.\ref{TSNE23} indicates that the proposed method is able to diagnose the fault of the pitch system of WT 23 excellently. The confusion matrix shown in Fig.\ref{confusion23} indicates that the proposed method can effectively solve the problem of difficulty in distinguishing multiple faults in the pitch system of WT 23.


\begin{table}[htb]
 \centering
 \caption{Comparison of fault diagnosis methods performance for WT 23}
 \label{resultWT23}
 \begin{tabular}{lccc}
  \toprule
  Algorithms & \text{Accuracy} & \text{macro G-mean} \\
  \midrule
  \textbf{HSMSCL}  & \textbf{0.9979} & \textbf{0.9991}\\
  SCL-MLP  & 0.9467 &  0.9693\\
  MLP+FL & 0.9759 & 0.8982 \\
  ResNet101+FL & 0.9728 & 0.9634\\
  BiLSTM+FL & 0.9129 & 0.9351\\
  CNN+FL & 0.9501 &  0.9509\\
  \bottomrule
 \end{tabular}
\end{table}

\begin{figure*}[htbp]
\centering
\subfloat[]{\includegraphics[width=2in]{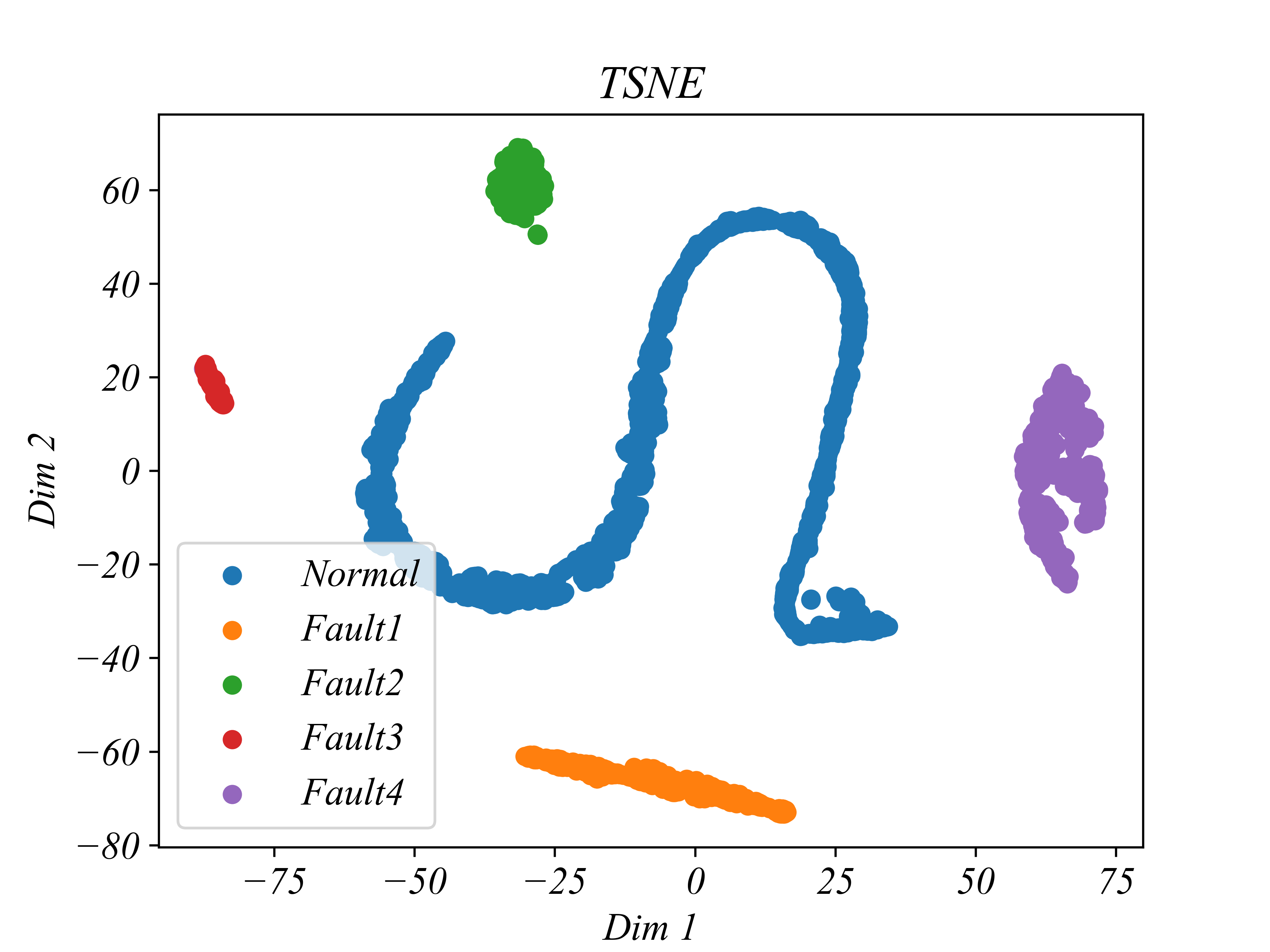}%
\label{TSNE29HSMSCL}}
\hfil
\subfloat[]{\includegraphics[width=2in]{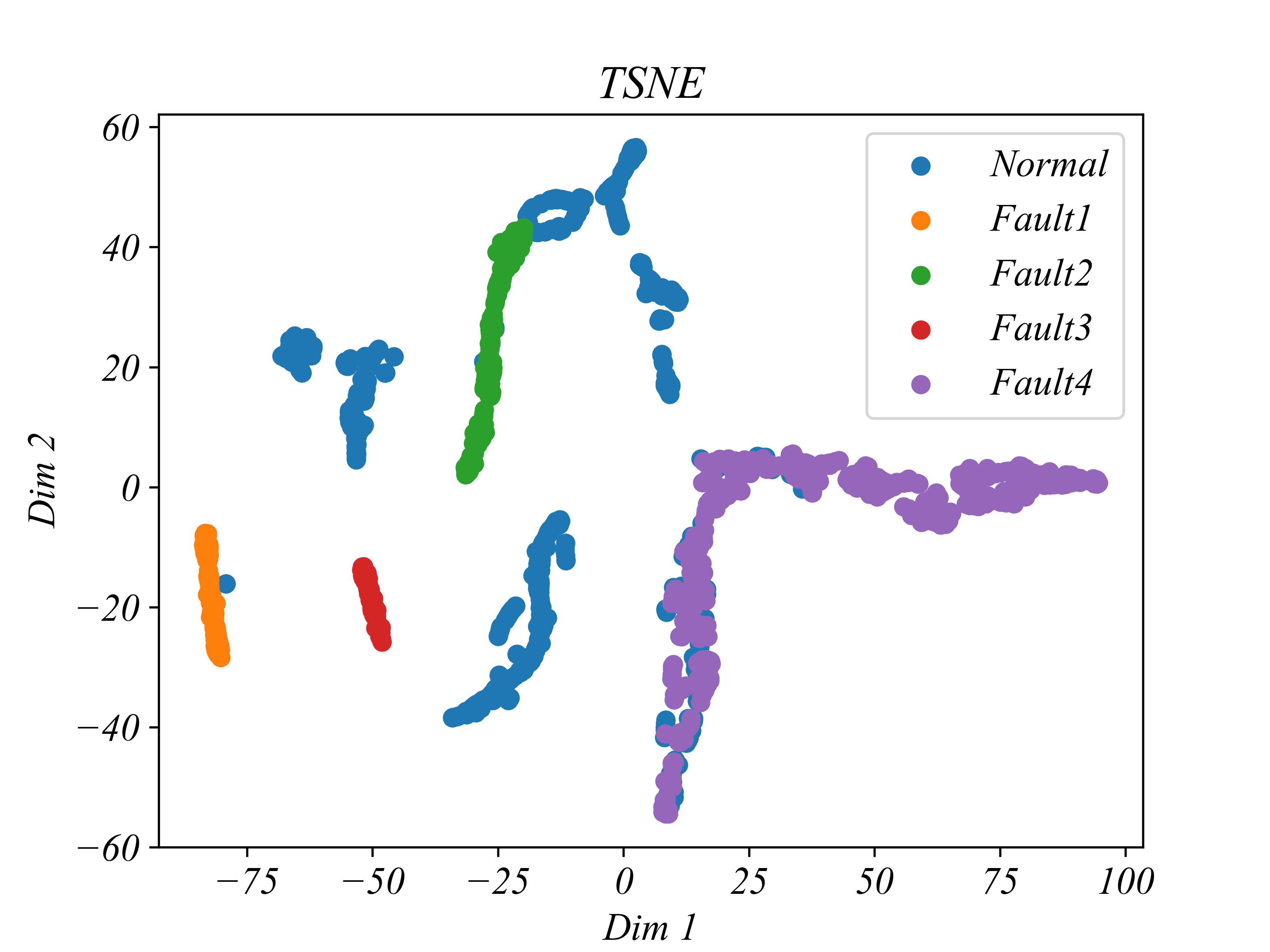}%
\label{TSNE29SCL}}
\hfil
\subfloat[]{\includegraphics[width=2in]{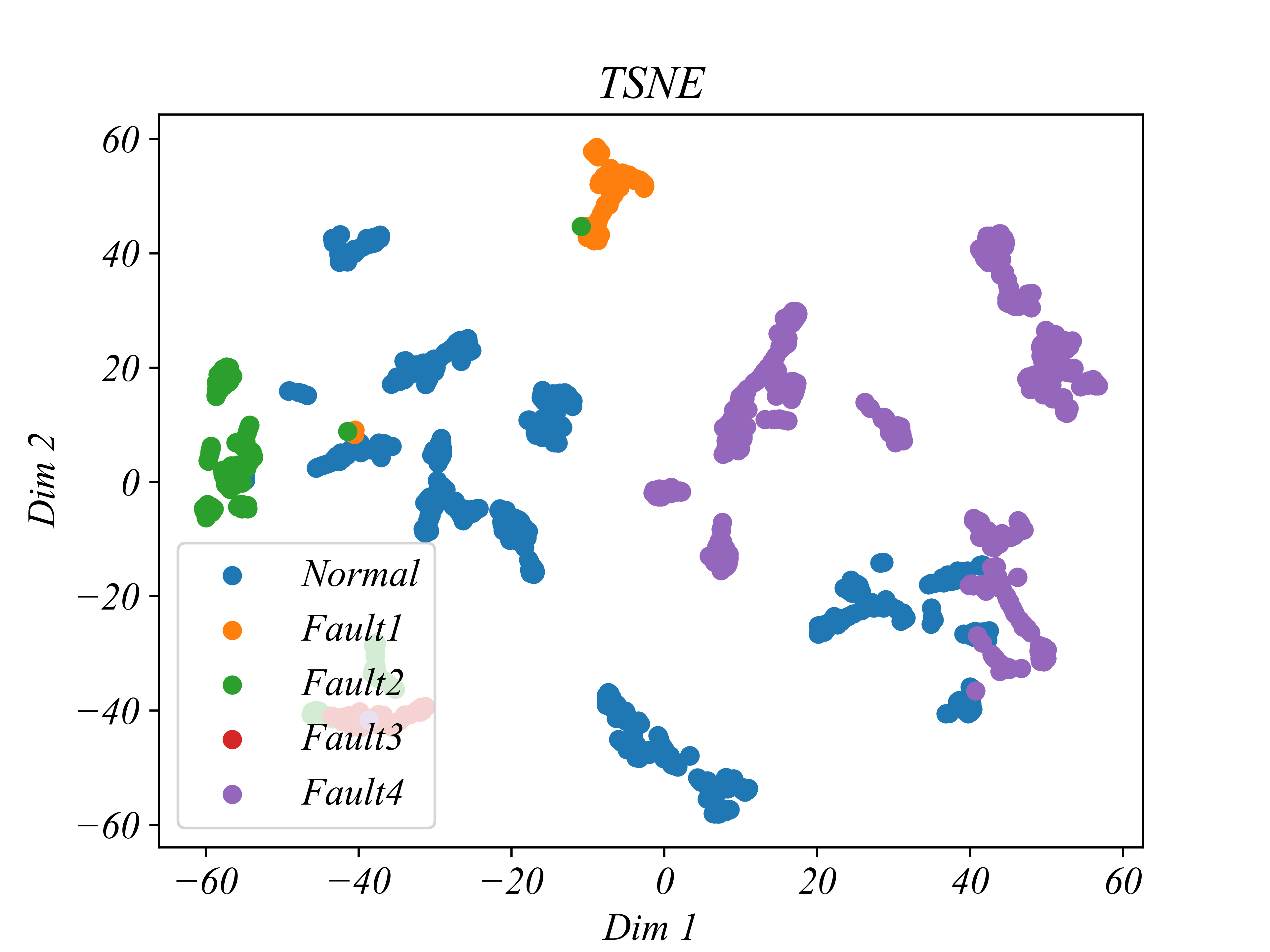}%
\label{TSNE29MLP}}
\hfil
\\
\subfloat[]{\includegraphics[width=2in]{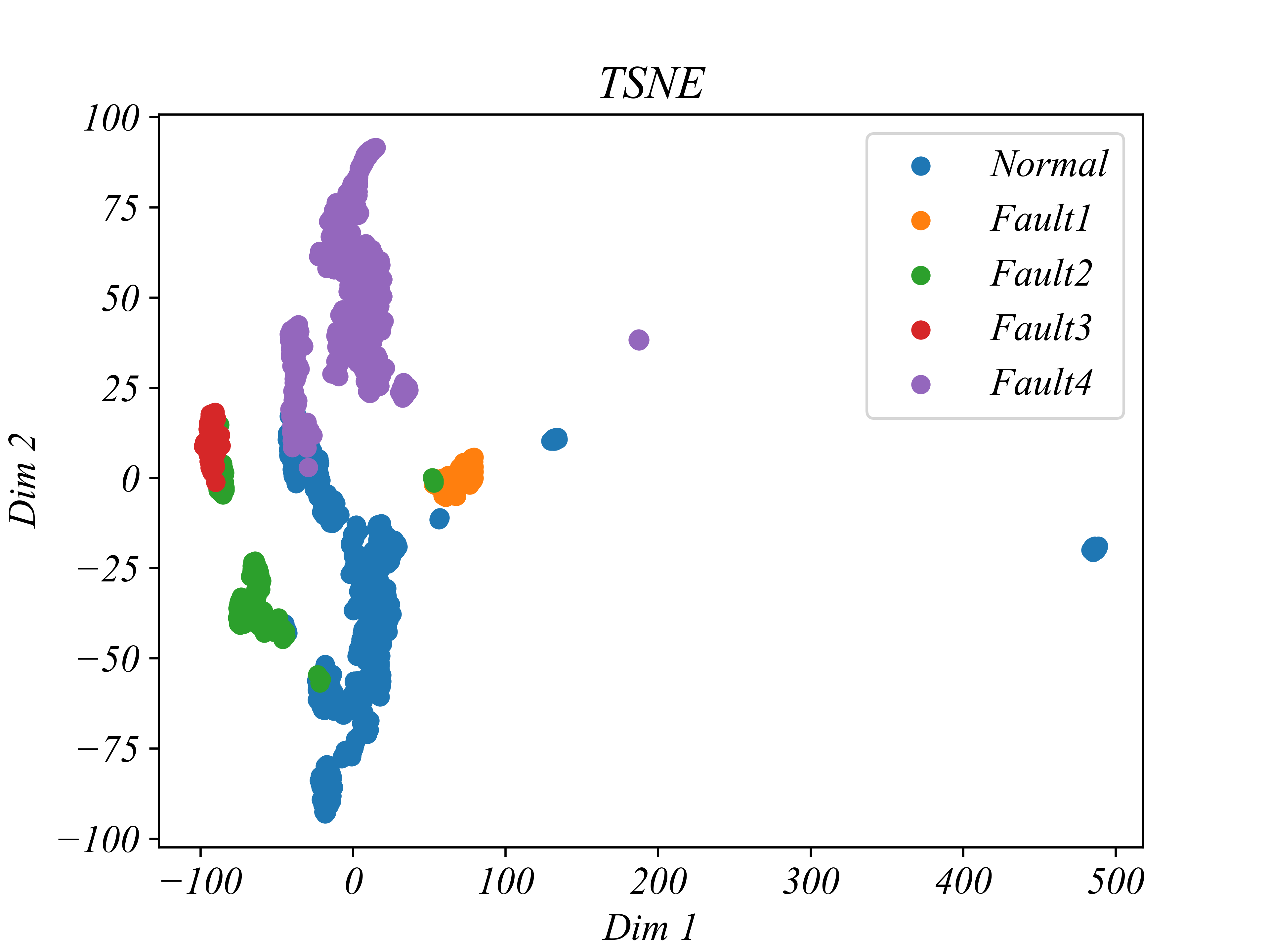}%
\label{TSNE29Resnet}}
\hfil
\subfloat[]{\includegraphics[width=2in]{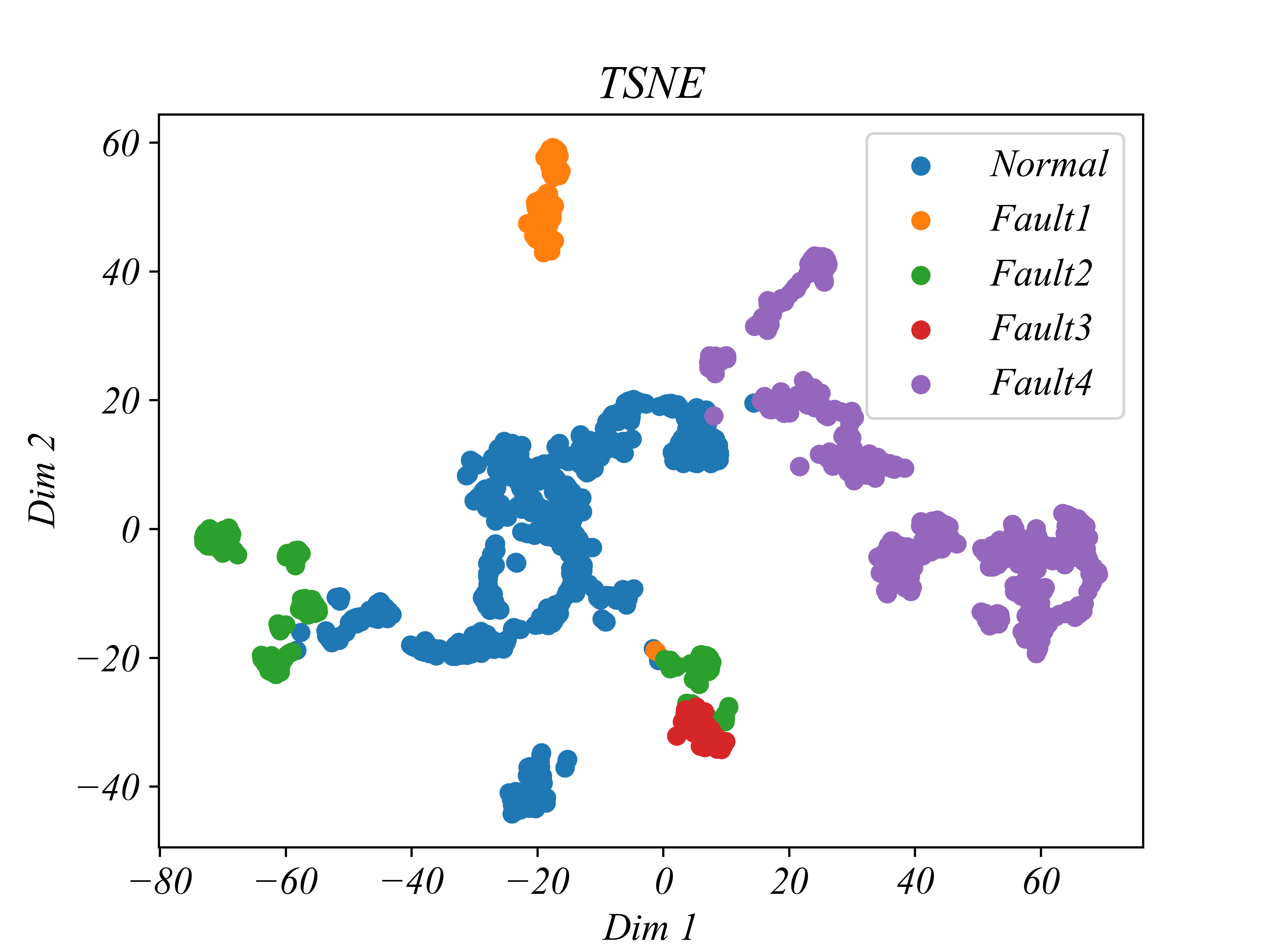}%
\label{TSNE29LSTM}}
\hfil
\subfloat[]{\includegraphics[width=2in]{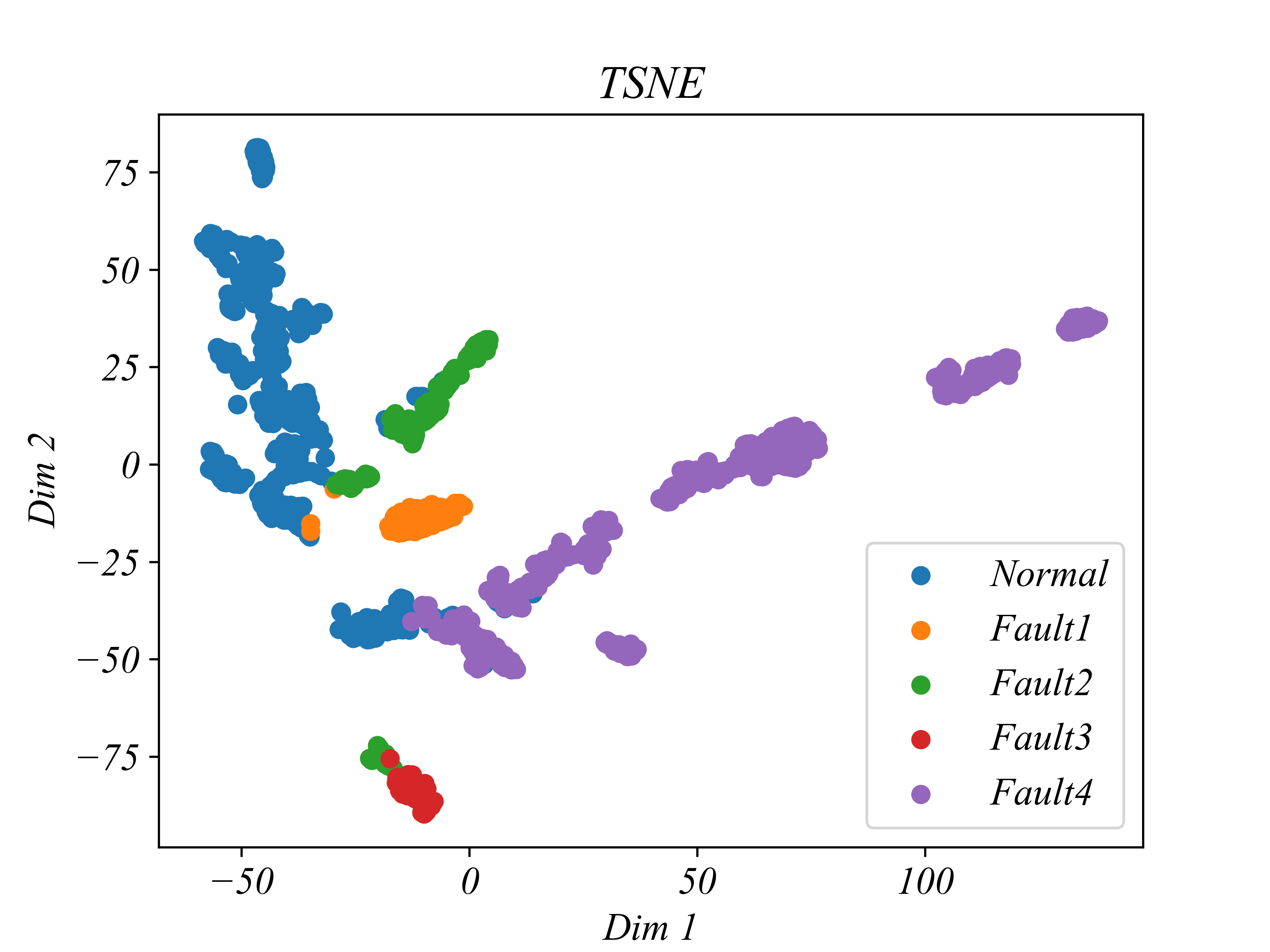}%
\label{TSNE29CNN}}
\caption{T-SNE embedding for WT 29 (a) Proposed method (b) SCL+MLP (c) MLP+FL (d) ResNet101+FL (e) BiLSTM+FL (f) CNN+FL}
\label{TSNE29}
\end{figure*}

\begin{figure*}[htbp]
\centering
\subfloat[]{\includegraphics[width=2in]{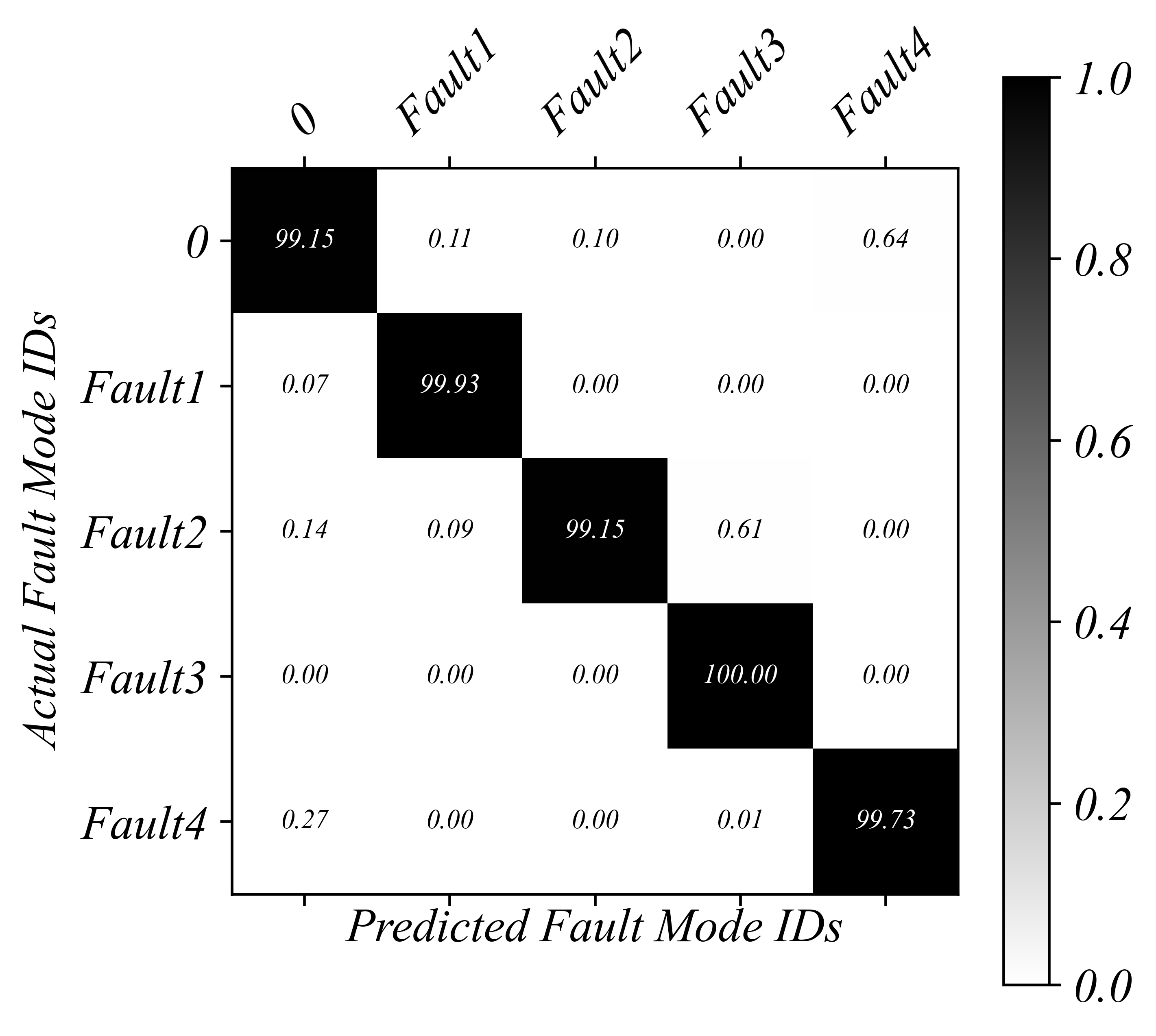}%
\label{confusion29HSMSCL}}
\hfil
\subfloat[]{\includegraphics[width=2in]{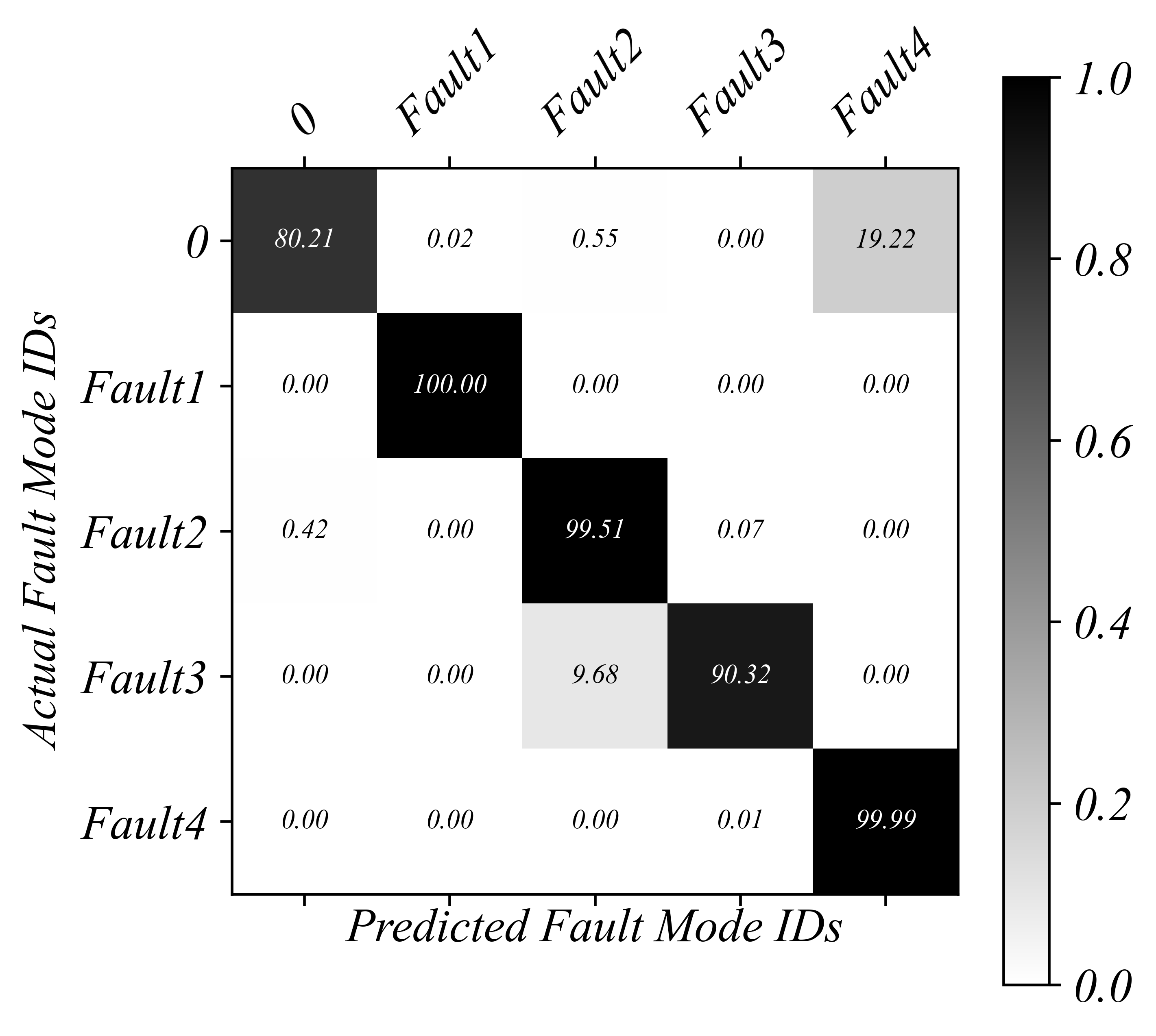}%
\label{confusion29SCL}}
\hfil
\subfloat[]{\includegraphics[width=2in]{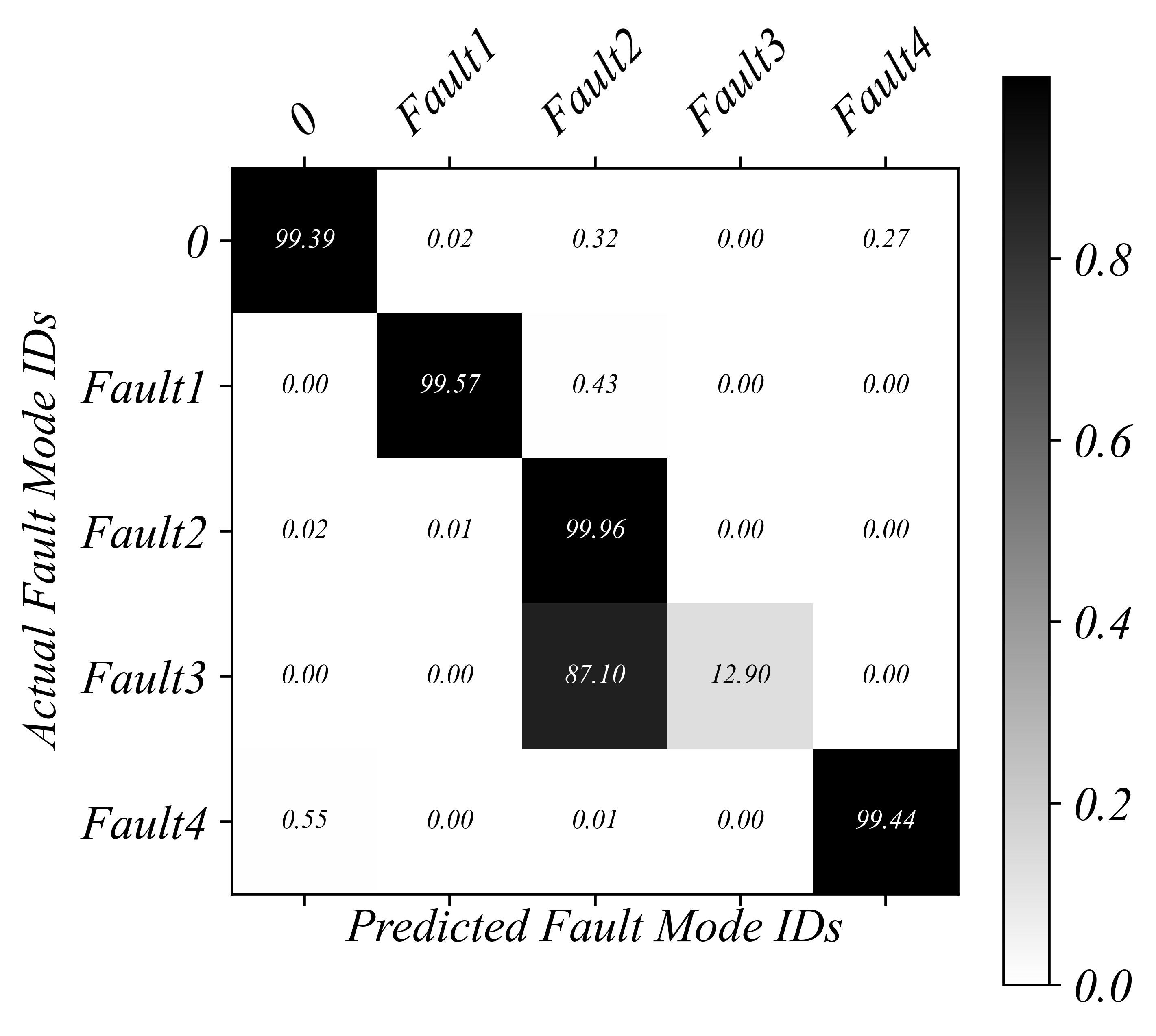}%
\label{confusion29MLP}}
\hfil
\\
\subfloat[]{\includegraphics[width=2in]{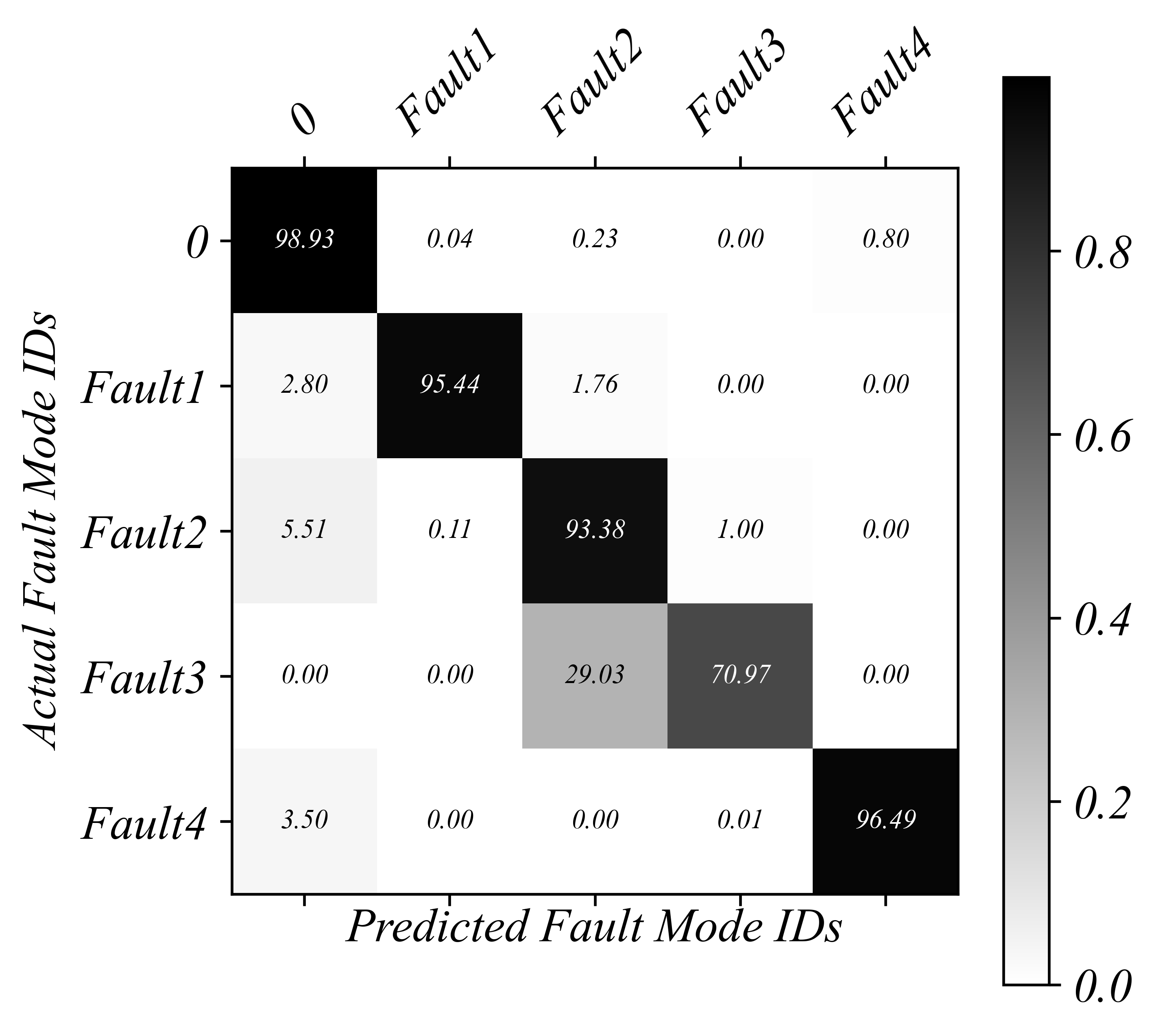}%
\label{confusion29Resnet}}
\hfil
\subfloat[]{\includegraphics[width=2in]{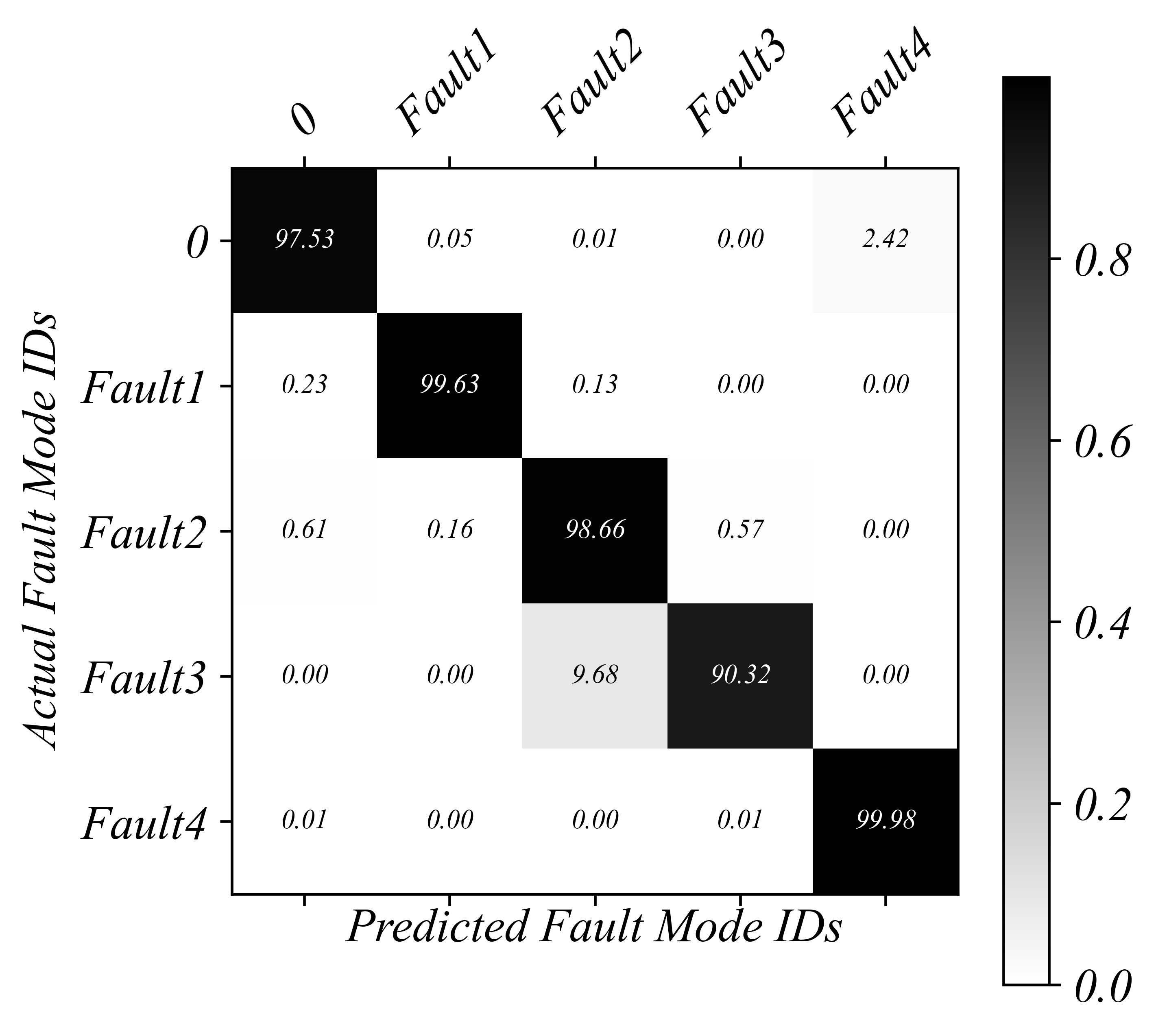}%
\label{confusion29LSTM}}
\hfil
\subfloat[]{\includegraphics[width=2in]{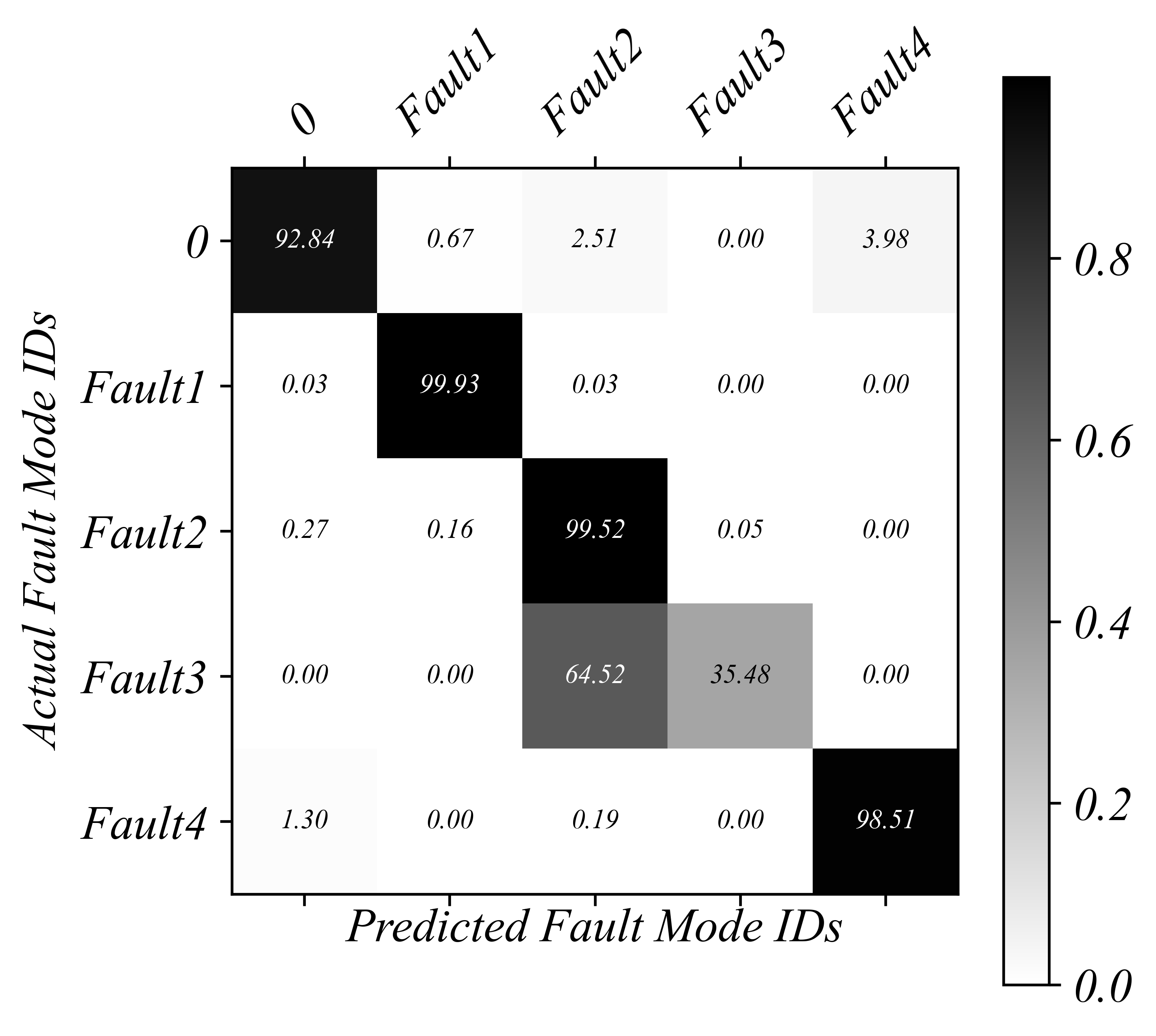}%
\label{confusion29CNN}}
\caption{Confusion matrix for WT 29 (a) Proposed method (b) SCL+MLP (c) MLP+FL (d) ResNet101+FL (e) BiLSTM+FL (f) CNN+FL}
\label{confusion29}
\end{figure*}

\subsubsection{Experiment for WT 29}

To verify the generalization performance of the proposed method, we also compared it with baselines on WT 29. We repeated the experiment ten times with different random seeds and the reported results are based on the average. The results are shown in Table \ref{resultWT29}. The corresponding t-SNE embeddings and confusion matrix are shown in Fig.\ref{TSNE29} and Fig.\ref{confusion29}.  Compared with the baselines, the \text{Accuracy} of the proposed method is increased by 9.46\%, 0.45\%, 4.97\%, 2.07\%, 1.68\% and 1.09\%, and the \text{macro G-mean} of the proposed method is increased by 4.39\%, 7.85\%, 3.01\%, 14.02\%, 2.31\% and 6.02\%. The results in Table \ref{resultWT29}, Fig. \ref{TSNE29} and Fig. \ref{confusion29} once again verify that the proposed method has excellent hard-sample classification capability, which means that the proposed method is applicable to the hard-to-classify problem present in the wind turbine pitch system.


\begin{table}[htb]
 \centering
 \caption{Comparison of fault diagnosis methods performance for WT 29}
 \label{resultWT29}
 \begin{tabular}{lccc}
  \toprule
  Algorithms & \text{Accuracy} & \text{macro G-mean} \\
  \midrule
  \textbf{HSMSCL}  & \textbf{0.9940} & \textbf{0.9971}\\
  SCL-MLP  & 0.9081 &  0.9552\\
  MLP+FL & 0.9945 & 0.9062 \\ 
  ResNet101+FL & 0.9719 & 0.9495\\
  BiLSTM+FL & 0.9867 & 0.9840\\
  CNN+FL & 0.9605 &  0.9180\\
  \bottomrule
 \end{tabular}
\end{table}

\subsection{Impact of hyperparameter temperature $\tau$}

\begin{figure}[htb]
\centering
\includegraphics[width=0.95\linewidth]{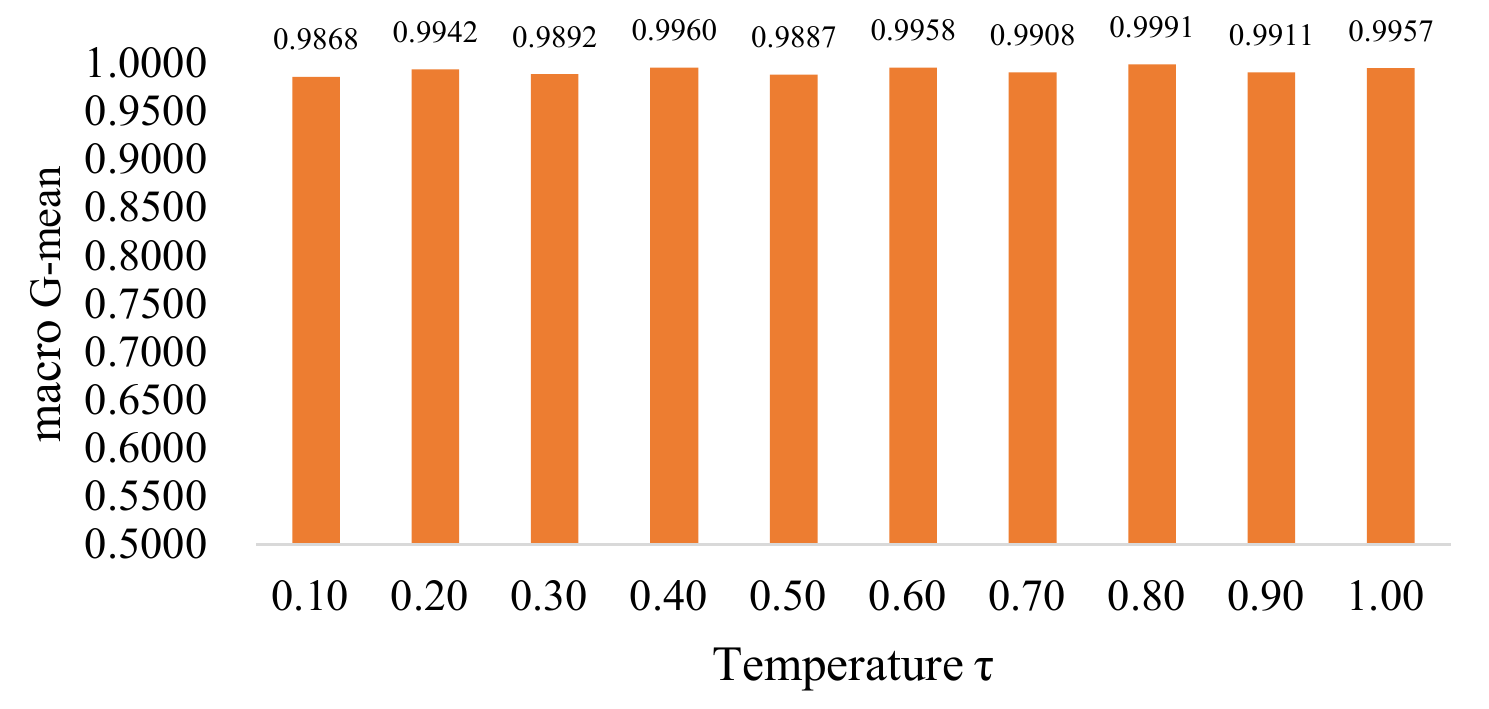}
\caption{macro G-mean as a function of the temperature parameter $\tau$}
\label{sensitivitytem}
\end{figure}

The hyperparameter temperature is commonly used in contrastive learning \cite{wang2021understanding}, we have performed experiments on the proposed method by selecting different temperature values including 0.1, 0.2, 0.3, 0.4, 0.5, 0.6, 0.7, 0.8, 0.9 and 1.0. As shown in Fig. \ref{sensitivitytem}, the proposed fault diagnosis method's performance is excellent at different temperature values, and the best performance is achieved when the temperature is 0.8. Therefore, the fault diagnosis performance of the proposed method is robust to different values of temperature $\tau$.

\begin{table}[htb]
 \centering
 \caption{Ablation analysis of HSM for WT 23}
 \label{ablation23}
 \begin{tabular}{lccc}
  \toprule
  Algorithms &  \text{macro G-mean} \\
  \midrule
  \textbf{HSMSCL}   & \textbf{0.9991}\\
  HSM+SCL-MLP  &   0.9887\\
  SCL-HSM+MLP &  0.9808\\
  \bottomrule
 \end{tabular}
\end{table}

\begin{table}[htb]
 \centering
 \caption{Ablation analysis of HSM for WT 29}
 \label{ablation29}
 \begin{tabular}{lccc}
  \toprule
  Algorithms &  \text{macro G-mean} \\
  \midrule
  \textbf{HSMSCL}  &  \textbf{0.9971}\\
  HSM+SCL-MLP  &   0.9682\\
  SCL-HSM+MLP &  0.9595\\
  \bottomrule
 \end{tabular}
\end{table}

\subsection{Ablation analysis of HSM}

To verify the effectiveness of the HSM framework for the two stages (SCL and MLP) in the proposed method, 
we compare the proposed method with the following two methods and evaluate their performance by \text{macro G-mean}.

HSM+SCL-MLP: The HSM framework is removed from MLP in the proposed method.

SCL-HSM+MLP: The HSM framework is removed from SCL in the proposed method.

The results of the ablation analysis are shown in Table \ref{ablation23} and Table \ref{ablation29}. From these two tables, it can be seen that \text{macro G-mean} of the proposed method is significantly higher than that of the two compared methods. Taking the experimental results on WT 23 shown in Table \ref{ablation23} as an example, the \text{macro G-mean} of the proposed method is 1.05\% and 1.87\% higher than that of HSM+SCL-MLP and SCL-HSM+MLP, respectively. The above results demonstrate the necessity of the HSM framework for the proposed method.

\section{Conclusion}

In this paper, hard sample mining enabled unified contrastive feature learning is proposed to diagnose the fault in wind turbine pitch systems. The existence of hard samples between different health conditions is a practical problem in wind turbine pitch systems. The proposed method solves this problem by combining a hard sample mining framework with a two-step approach consisting of supervised contrastive learning and multi-layer perceptron. 
Notably, few previous works have focused on the problem of hard samples in the field of wind turbine fault diagnosis, with this paper being a pioneer in applying hard sample mining techniques to this field. The effectiveness of the proposed method is validated on two real wind turbine pitch system failure datasets. As demonstrated in the experiments, the proposed HSMSCL outperforms the baselines. This means that the proposed method introduces a novel and effective solution for wind turbine pitch system fault diagnosis.

\bibliographystyle{unsrtnat}
\bibliography{ref.bib}

\newpage

\vfill

\end{document}